\documentclass[journal]{IEEEtran}

\usepackage[utf8]{inputenc}
\usepackage{xcolor}
\usepackage{amsmath}
\usepackage{graphicx}
\usepackage{times}
\usepackage{soul}
\usepackage[small]{caption}
\usepackage{amsfonts}
\usepackage{array}
\usepackage{multirow}
\usepackage{longtable}
\usepackage{subfigure}
\usepackage{algorithm}
\usepackage{algorithmicx}
\usepackage{algpseudocode}
\usepackage{cite}
\usepackage{pifont}% http://ctan.org/pkg/pifont
\usepackage{arydshln} % for cdashline
\usepackage{booktabs}
\usepackage[normalem]{ulem}
\usepackage{soul}
\newcommand{\cmark}{\ding{51}}%
\newcommand{\xmark}{\ding{55}}%
\usepackage{csquotes}

\algnewcommand\algorithmicinput{\textbf{Input:}}
\algnewcommand\INPUT{\item[\algorithmicinput]}
\algnewcommand\algorithmicoutput{\textbf{Output:}}
\algnewcommand\OUTPUT{\item[\algorithmicoutput]}

\def\eg{\emph{e.g.}} 
\def\ie{\emph{i.e.}} 
 
 \def\vs{\emph{vs}}

\def\vs{\emph{v.s.}}

\def\RR{\textcolor{black}}

\def\RRR{\textcolor{black}}
\def\RRRR{\textcolor{black}}

\usepackage{amsmath,amsfonts,bm}

\def\Tabref#1{Table~\ref{#1}}

\def\Figref#1{Fig.~\ref{#1}}

\def\Secref#1{Section~\ref{#1}}
\def\eqref#1{equation~\ref{#1}}
\def\Eqref#1{Eq.~\ref{#1}}
\def\1{\bm{1}}
\newcommand{\train}{\mathcal{D}}

\def\rvx{{\mathbf{x}}}
\def\rvy{{\mathbf{y}}}

\def\rmM{{\mathbf{M}}}

\DeclareMathAlphabet{\mathsfit}{\encodingdefault}{\sfdefault}{m}{sl}
\SetMathAlphabet{\mathsfit}{bold}{\encodingdefault}{\sfdefault}{bx}{n}

\def\gD{{\mathcal{D}}}

\def\gF{{\mathcal{F}}}

\def\gY{{\mathcal{Y}}}

\newcommand{\R}{\mathbb{R}}

\DeclareMathOperator*{\argmin}{arg\,min}

\begin{document}
%
% paper title
% Titles are generally capitalized except for words such as a, an, and, as,
% at, but, by, for, in, nor, of, on, or, the, to and up, which are usually
% not capitalized unless they are the first or last word of the title.
% Linebreaks \\ can be used within to get better formatting as desired.
% Do not put math or special symbols in the title.

% \title{Meta Attribute Based Filter Pruning \\ for Efficient Deep Convolutional Neural Networks}

\title{\RR{Filter Pruning by Switching to} \\ \RR{Neighboring CNNs with Good Attributes}}
%Progressive Acceleration Efficient inference

\author{Yang~He,
        Ping~Liu,
        Linchao~Zhu,
        and~Yi~Yang
%\author{Yang~He,~\IEEEmembership{Member,~IEEE,}
%        John~Doe,~\IEEEmembership{Fellow,~OSA,}
%        and~Jane~Doe,~\IEEEmembership{Life~Fellow,~IEEE}% <-this % stops a space
\thanks{Y. He, P.~Liu and L.~Zhu are with the ReLER lab, Australian Artificial Intelligence Institute, University of Technology Sydney, Sydney, NSW 2007, Australia (e-mail: yang.he-1@student.uts.edu.au; pino.pingliu@gmail.com; linchao.zhu@uts.edu.au).
Y. He and P. Liu are also with A*STAR Centre for Frontier AI Research (CFAR), Singapore 138632.
Y.~Yang is with the College of Computer Science and Technology, Zhejiang University, Hangzhou, China, 310000 (e-mail: yangyics@zju.edu.cn).
L.~Zhu is the corresponding author.}% <-this % stops a space
%\thanks{Y. He and Y.~Yang are also with the SUSTech-UTS Joint Centre of CIS (SUCCIS), Southern University of Science and Technology, Guangdong 518005, China.}% <-this % stops a space
%\thanks{Manuscript received April 19, 2005; revised August 26, 2015.}
}

% The paper headers
\markboth{Journal of \LaTeX\ Class Files,~Vol.~14, No.~8, August~2015}%
{Shell \MakeLowercase{\textit{et al.}}: Bare Demo of IEEEtran.cls for IEEE Journals}
% The only time the second header will appear is for the odd numbered pages
% after the title page when using the twoside option.
% 
% *** Note that you probably will NOT want to include the author's ***
% *** name in the headers of peer review papers.                   ***
% You can use \ifCLASSOPTIONpeerreview for conditional compilation here if
% you desire.

% If you want to put a publisher's ID mark on the page you can do it like
% this:
%\IEEEpubid{0000--0000/00\$00.00~\copyright~2015 IEEE}
% Remember, if you use this you must call \IEEEpubidadjcol in the second
% column for its text to clear the IEEEpubid mark.

% use for special paper notices
%\IEEEspecialpapernotice{(Invited Paper)}

% make the title area
\maketitle
% As a general rule, do not put math, special symbols or citations
% in the abstract or keywords.

\begin{abstract}
Filter pruning is effective to reduce the computational costs of neural networks.
Existing methods show that updating the previous pruned filter would enable large model capacity and achieve better performance.
However, during the iterative pruning process, even if the network weights are updated to new values, the pruning criterion remains the same.
In addition, when evaluating the filter importance, only the magnitude information of the filters is considered.
% As the filters of the network work jointly, merely magnitude information is not enough to reflect the property of the network.
\RR{However, in neural networks, filters do not work individually, but they would affect other filters. As a result, the magnitude information of each filter, which merely reflects the information of an individual filter itself, is not enough to judge the filter importance.
}
To solve the above problems, we propose Meta-attribute-based Filter Pruning (MFP).
First, to expand the existing magnitude information based pruning criteria, \RR{we introduce a new set of criteria to consider the geometric distance of filters.}
Additionally, to explicitly assess the current state of the network, we adaptively select the most suitable criteria for pruning via a meta-attribute, a property of the neural network at the current state.
Experiments on two image classification benchmarks validate our method. For ResNet-50 on ILSVRC-2012, we could reduce more than 50\% FLOPs with only 0.44\% top-5 accuracy loss. 
\end{abstract}

% Note that keywords are not normally used for peerreview papers.
\begin{IEEEkeywords}
Neural Networks, Filter Pruning, Network Compression, Meta-attributes
\end{IEEEkeywords}

% For peer review papers, you can put extra information on the cover
% page as needed:
% \ifCLASSOPTIONpeerreview
% \begin{center} \bfseries EDICS Category: 3-BBND \end{center}
% \fi
%
% For peerreview papers, this IEEEtran command inserts a page break and
% creates the second title. It will be ignored for other modes.
\IEEEpeerreviewmaketitle

\section{Introduction}
% The very first letter is a 2 line initial drop letter followed
% by the rest of the first word in caps.
% 
% form to use if the first word consists of a single letter:
% \IEEEPARstart{A}{demo} file is ....
% 
% form to use if you need the single drop letter followed by
% normal text (unknown if ever used by the IEEE):
% \IEEEPARstart{A}{}demo file is ....
% 
% Some journals put the first two words in caps:
% \IEEEPARstart{T}{his demo} file is ....
% 
% Here we have the typical use of a "T" for an initial drop letter
% and "HIS" in caps to complete the first word.

\IEEEPARstart{T}{he} computational cost of deep convolutional neural networks (CNNs) for different computer vision tasks~\cite{frenay2013classification,yan2016image,gong2015change,zhang2020sgone,yang2012feature,zhu2021temporal} is always on the rise due to the complex architectures of modern CNNs such as VGGNet~\cite{simonyan2014very} and ResNet~\cite{he2016deep}. To deploy those complicated models on low resource devices, network pruning is essential since it reduces the storage demand and computational costs (knows as floating point operations or FLOPs) of neural networks.
To accelerate the networks and reduce the model size, researchers propose weight pruning~\cite{han2015learning,carreira2018learning}, and the filter pruning~\cite{li2016pruning,yu2018nisp}. In recent works, filter pruning is the preferred method. \RR{This preference stems from the fact that filter pruning could remove the whole filters, creating a model with structured sparsity.} With the structured sparsity, the model could take full advantage of high-efficiency Basic Linear Algebra Subprograms (BLAS) libraries to achieve better acceleration.

\begin{figure}[ht]
\center
\includegraphics[width=0.99\linewidth]{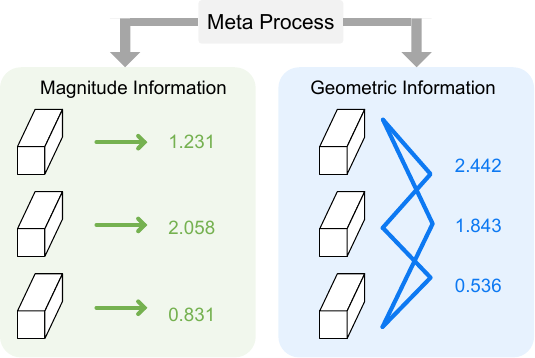}
\caption{
The meta-process selects suitable information for pruning. \RR{Magnitude information depends on the filter itself, while geometric information is based on two filters.}
}
\label{fig:overview}
\end{figure}

Recently, pruning filters in a \textit{soft manner}~\cite{guo2016dynamic,he2018soft,chen2019layer,lin2016towards,wang2019one,dai2018grow} have proven to be more effective than the conventional three-stage pipeline.
The advantage of soft filter pruning is that previously pruned filters can be updated during training. \RR{Once the training process is complete, the removal of unimportant filters from the network can take place.} In this way, the model’s capacity will not decrease, and network performance is enhanced.

% The advantage of soft manner means that after pruning, the previous pruned filters could still be updated during training.
% Only when the training process is completely finished, the unimportant filters are truly pruned from the network.
% In this way, the model capacity would not decrease, and the performance of the network is better.
%the weakness of the conventional three-stage pipeline, that is, during training, would not happen for methods with \textit{soft manner}.

However, previous works using the \textit{soft filter pruning} have two drawbacks.
\RR{First, previous filter pruning works often ignore the geometric information between filters when pruning.}
Prevailing filter pruning works~\cite{li2016pruning,ye2018rethinking,he2018soft} are based on the ``smaller-norm-less-important" criterion. 
The basis of this criterion is that filters with smaller norms are less critical, and therefore, pruning those filters with smaller $\ell_{1}$-norm~\cite{li2016pruning} or $\ell_{2}$-norm~\cite{he2018soft} will not bring dramatic performance drop. 
\RR{ 
Although the $\ell_{p}$-norm-based criterion focuses on the magnitude information of individual filters, the criterion ignores the geometric information between filters; a disadvantage that may lead to inappropriate pruning results.
% However, one disadvantage of the $\ell_{p}$-norm based criterion is that it focuses on the magnitude information of individual filters but ignores the geometric information between them, and thus might lead to inappropriate pruning results.
Research by \cite{yosinski2015understanding} show that the filters in the network have regular patterns, thus considering the geometric information is critical for pruning.}
Here is an example, suppose we have three filters to prune, each of which is a three-dimension vector: $A=(1,1,1)$, $B = (1.1, 1, 1)$, and $C = (0.5, 0.3, 0.2)$. Norm-based criterion would prune $C$ since it has the smallest norm. However, if we look closely at $A$ and $B$, we will find that they are statistically similar. \RR{Therefore, they make a very similar, if not the same, contribution to the network, which means pruning either $A$ or $B$ is more reasonable.}
\RRRR{We need to be careful when the value of a filter becomes zero. Filter $C$ will become zero only when the filter is \textbf{pruned}. Under this condition, filter $C$ should remain zero since the pruned filters should not change to a non-zero value.}
Second, in previous works~\cite{guo2016dynamic,he2018soft,chen2019layer,lin2016towards,wang2019one,dai2018grow}, the weights of the network are updated to new values after pruning and training.
However, the pruning criterion during the entire training and pruning process keeps fixed and fails to adapt to current conditions.
A fixed criterion may not be able to evaluate a network of different weight values.
%   does not change accordingly
\RRRR{For the first problem, we introduce new measures to rank filters, using their geometric information.} 
Assume a filter space contains all the filters of the network, different weights values indicate that the filters are located in various positions of the filter space.
\RR{In this way, we could utilize the distance between the filters to characterize the geometric information among filters.
Specifically, Minkowski distance, an effective similarity measure \cite{singh2013k}, is employed to measure the geometric information.
After we obtain the geometric information of filters, we can prune the filters that make a replaceable contribution to the network.}
To solve the second problem, we build a Meta-attribute-based Filter Pruning (MFP) framework to adaptively select the most appropriate criteria based on the current state of the network. 
As shown in Fig.~\ref{fig:overview}, during pruning and training, networks with different values would have various meta-attributes~\cite{brazdil2003ranking}. \RR{These meta-attributes are particularly effective in measuring the difference between the networks and assist in choosing the optimally pruned one from the candidates.}
Inspired by the meta-learning framework~\cite{brazdil2003ranking}, we minimize the meta-attributes of the pruned model and the original model, which enables the pruned models to perform as well as the original models on the pre-defined task.  
% \RR{For the first problem, we introduce new measurements to model geometric information among filters.} \hl{For the first problem, we introduce new measurements to model co-relation among filters by utilizing their geometric information.}
% Assume a filter space contains all the filters of the network. Different weights filter values indicate that the filters are located in different positions of the filter space.
% \RR{In this way, we could utilize the distance between the filters to characterize the geometric information among filters.
% Specifically, Minkowski distance is employed to measure the geometric information as it is an effective similarity measure \cite{singh2013k}.
% After we obtain the geometric information of filters, we can prune the filters that make a replaceable contribution to the network.}
% To solve the second problem, we build a Meta-attribute-based Filter Pruning (MFP) framework to adaptively select the most suitable criteria based on the current state of the network. 
% As shown in Fig.~\ref{fig:overview}, during pruning and training, network with different values would have various meta-attributes~\cite{brazdil2003ranking}, which is metric to measure the difference of the network.
% Inspired from the meta-learning framework~\cite{brazdil2003ranking}, we minimize the meta-attributes of the pruned model and the original model, by doing which we make the pruned models perform as well as the original models on the pre-defined task.

\textbf{Contributions.} We outline three contributions:

(1)~\RRRR{We introduce distance-based measures to model the geometric  information among the filters, which is complementary to the existing magnitude measures.}
% \hl{The introduced distance-based measures consider the co-relation between filters in pruning,  and treat them as a cooperative group rather than independent individuals.}
% (1)~\RR{We introduce distance-based measurements to model the geometric  information among the filters, which is complementary to the existing magnitude measurements.}

(2) We utilize the meta-attribute to characterize the difference between pruned and original models.
Following a \textit{meta-learning} way, we adaptively select a suitable meta-attribute based on the current network state and conduct pruning based on the selected meta-attribute.

(3)~The experiments on two benchmarks validate the effectiveness of our MFP.
For ResNet-50 on ILSVRC-2012, we could reduce more than 50\% FLOPs with only 0.44\% top-5 accuracy loss.
Moreover, our MFP accelerates ResNet-110 on CIFAR-10 by two times with even 0.16\% relative accuracy improvement.
%-------------------------------------------------------------------------
\begin{figure*}[!ht]
\center
\includegraphics[width=0.99\linewidth]{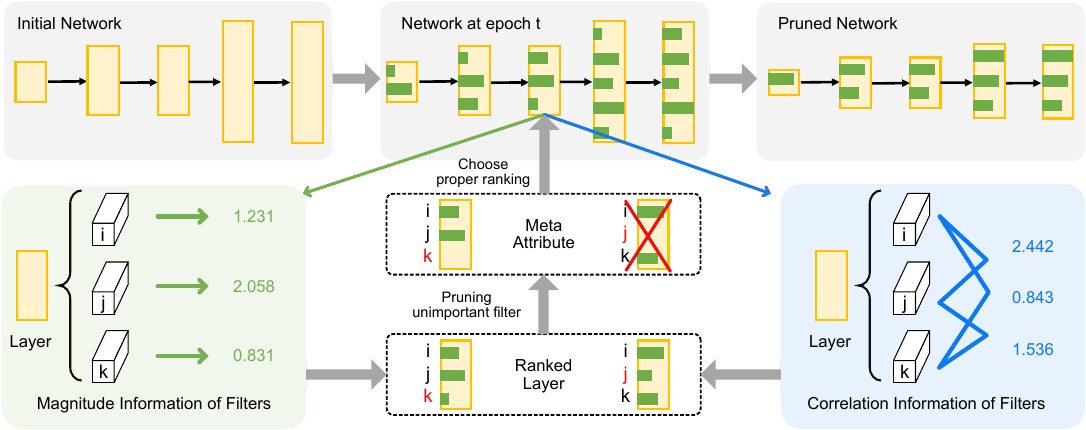}
\caption{
\RR{The process of our pruning method. Filters of a layer would be ranked based on the magnitude and the geometric information.} Then the meta-attribute of the network is utilized to determine which information should be used for pruning at the current state.
}
\label{fig:process}
\end{figure*}

\section{Related Work}

Based on the granularity of pruning, network pruning can be divided into weight pruning~\cite{han2015learning,han2015deep,guo2016dynamic,tung2018clip,carreira2018learning,zhang2018systematic,dong2017learning} and filter pruning~\cite{li2016pruning}. The former one focuses on pruning the fine-grained weight of filters, but the unstructured sparsity in the pruned model makes it less friendly for realistic acceleration.
On the other hand, filter pruning achieves structured sparsity and a realistic acceleration with high-efficiency Basic Linear Algebra Subprograms (BLAS) libraries.

\subsection{Weight Pruning.}

Many recent works~\cite{han2015learning,han2015deep,guo2016dynamic} prune weights of the neural network to produce small models. For example,
\cite{han2015learning} proposes an iterative weight pruning method by discarding the small weights with values below the threshold.
Dynamic network surgery is employed by \cite{guo2016dynamic} to reduce the training iteration while maintaining good prediction accuracy.
%\cite{guo2016dynamic} proposed
%dynamic network surgery which is a network compression method.
\cite{wen2016learning,lebedev2016fast} leverage the sparsity property of feature maps or weight parameters to accelerate the CNN models.
A special case of weight pruning is known as neuron pruning.
\cite{hu2016network} evaluates the importance of neurons by measuring the sparsity of ReLU activations.
% percentage of zero activations of a neuron after the ReLU mapping.
% To this end, \cite{wen2016learning} proposed the Structured Sparsity
% Learning (SSL) method to regularize filter, channel, filter shape
% and depth structures. \cite{lebedev2016fast} applied the group-sparsity
% regularization on the loss function to shrink some entire groups of
% weights towards zeros.
% Despite their success in \cite{wen2016learning,lebedev2016fast},
% it is still computational expensive in computing gradients of the additional regularization term with respect to all the weights.
% Neuron Pruning has also been considered as 
% The neurons with lower importance can be pruned.
However, pruning weights always leads to unstructured models, so the model cannot leverage the existing efficient BLAS libraries in practice. Therefore, it is difficult for weight pruning to achieve a realistic speedup.
% and thus inefficient in reducing the inference time of the network.

\subsection{Filter Pruning.}

If we consider whether to utilize the training data to determine the pruned filters, we could further divide the filter pruning methods into two categories:

\noindent 
{\bf Training Data Dependent Filter Pruning.}
~\cite{Liu_2017_ICCV,Luo_2017_ICCV,He_2017_ICCV,molchanov2016pruning,dubey2018coreset,suau2018principal,yu2018nisp,wang2018exploring,zhuang2018discrimination,huang2018learning,he2018adc,lin2019toward,chen2020dynamical,wang2017novel,lin2021filter,lin2021network,lin2019toward,lin2018holistic} utilize the training data to determine the pruned filters.
% \cite{suau2018principal} adopts the Principal Component Analysis (PCA) method to specify which part of the network should be preserved.
% \cite{Luo_2017_ICCV} proposes to use the information from the next layer to guide the filter selection.
% \cite{dubey2018coreset} minimizes the reconstruction error of training set sample activations and applies Singular Value Decomposition (SVD) to obtain a decomposition of filters.
% \cite{wang2018exploring} explores the linear relationship identified in different feature maps to eliminate the redundancy in convolutional filters.
\cite{he2018adc} leverages the reinforcement learning to find the redundancy for each layer automatically.
Sparsity regularization on the scaling factors of the network is imposed by \cite{Liu_2017_ICCV}.
He et al. \cite{He_2017_ICCV} utilizes the LASSO regression to select channels.
Yu et al. \cite{yu2018nisp} proposes to minimize the reconstruction error of important responses in the “final response layer”, and derives a closed-form solution to prune neurons in earlier layers.
\RRR{Kang et al.~\cite{kang2020operation} proposes data-driven SCP algorithm to prune models in a differentiable way.} 
\RRR{Gao et al.~\cite{gao2021network} considers the loss-metric mismatch problem and uses NPPM to solve the problem in pruning.} 
\RRR{DSA~\cite{ning2020dsa} finds the layer-wise pruning ratios with gradient-based optimization.}

\noindent
{\bf Training Data Independent Filter Pruning.}
Some training data-independent filter pruning methods~\cite{li2016pruning,he2018soft,ye2018rethinking,zhuo2018scsp,he2019filter,he2019asymptotic,he2020learning,he2019meta,wang2020progressive,wang2021soft} have been proposed. 
\cite{li2016pruning} and \cite{he2018soft} prune the filters with the $\ell_{1}$-norm criterion and $\ell_{2}$-norm criterion, respectively.
Research by \cite{he2018pruning} claims that the filters near the geometric median should be pruned, \cite{ye2018rethinking} proposes to prune the network by introducing sparsity on the scaling parameters of batch normalization (BN) layers,
%It forces the output of some channels being constant during training and uses an adaptation of the ISTA algorithm to update the batch-norm parameter. 
and \cite{zhuo2018scsp} clusters the filters in the spectral domain to select the unimportant filters.
% It is more practical than the first one as the training data may not be available during pruning operations.
\RRR{CLR~\cite{le2021network} finds a better learning rate schedule for pruning.}

\RRRR{Some previous work~\cite{sen2009meta,dong2017learning,huang2018learning,chin2018layer,liu2019metapruning} use the ``meta'' word or a similar term such as ``learning to prune'', but the core idea is different from our proposed mete-attribute-based filter pruning method. }
\cite{sen2009meta} focuses on modeling the complex software systems at high-levels of abstraction, rather than the neural network. In this situation, pruning refers to the removal of unnecessary classes and properties of software systems, not filters of the neural network.
\cite{dong2017learning} conducts pruning based on second-order derivatives of a layer-wise error function. The ``learning'' word means second order derivatives.
Moreover, \cite{huang2018learning} utilizes a reinforcement learning framework to determine the pruning filters.
~\cite{liu2019metapruning} generates the weights of the pruned network by a proposed PruningNet. To the best of our knowledge, none of the previous works mentions the meta-attributes for measuring the similarity of the original model and pruned model.

\RR{
Some of the previous works~\cite{wang2019cop} utilize the correlation for filter pruning, but there are some differences between their work and ours. 
First, \cite{wang2019cop} needs to calculate the pair-wise node-correlation first before calculating the filter correlation. In contrast, our approach  does not need extra node-correlation since we can directly calculate the distance between filters. \RRRR{Second, \cite{wang2019cop} adopts normalized correlation-based importance to measure the redundancy between two filters, while we use the Minkowski distance as a measure to include the filter scale.} Third, \cite{wang2019cop} contains two regularization terms in the final importance, which would introduce extra hyper-parameters. In contrast, our method does not rely on any regularization term since the filter distance can be used as the filter importance in a direct manner.
% First, before calculating the filter correlation, \cite{wang2019cop} needs to calculate the pair-wise node-correlation first. In contrast, our method could directly calculate the relation between filters and do not need extra node-correlation. Second, \cite{wang2019cop} uses normalized correlation-based importance to measure the redundancy between two filters, while we use the Minkowski distance as our measurement. Third, in \cite{wang2019cop}, the final importance of filters contains two regularization terms, which would introduce extra hyper-parameters. On the contrary, our method does not rely on any regularization term.
}

\RRRR{
The differences between Imp~\cite{molchanov2019importance} and our method include: 1) Different importance measure methods: Imp~\cite{molchanov2019importance} use Hessian matrix, while we use magnitude information and correlation information of filters. 2) Different dependence
 of dataset: Imp~\cite{molchanov2019importance} needs a few batches of input data to choose the filters, but we rely only on the weight values of the filters.
3) Different ways to select models: Imp~\cite{molchanov2019importance} utilizes importance score accumulation to select neurons, 
but we use meta-attributes to choose the pruned models.
}

\subsection{Neural Architecture Search}

\RRRR{
Some previous works use Neural Architecture Search (NAS) to automatically obtain the neural networks.
There are several differences between NAS and pruning. 1) Different objectives: NAS aims at obtaining new networks automatically, while pruning focus on compressing and accelerating the existing models to achieve higher performance. 2) Different components: NAS usually has many different kinds of cells as the searching components, while pruning methods consider the convolutional layers as the basic components. 
3) Different computations are required. Due to the large search space, NAS usually requires much more computational resources than pruning. While our method only costs several GPU days or even several GPU hours. 
DARTS~\cite{liu2018darts} proposes to search the architecture automatically using cells including separable convolutions, dilated separable convolutions, max pooling.
EfficientNet~\cite{tan2019efficientnet} searches the depth, width, and resolution of the network with mobile inverted bottleneck MBConv.
NASNet~\cite{zoph2018learning} requires 2000 GPU days for searching the architectures. 
% ENAS~\cite{pham2018efficient} proposes to share parameters between child models so the search process is fast. Note that the sub-architectures in DARTS and child models in ENAS need to be trained during searching, while our sub-networks do not. Our method is also different from one-shot
% NAS~\cite{brock2017smash} which generates the weights for sampled architecture using a HyperNet.
}

 %%%%%%%%%%%%%%%%%%%%%%%%%%%%%%%%%%%%%%%%%%%%%%%%%%%%%%%%%%%%%%%%%%%%%%%%%%%%%%%%%%%%%%%%%%%%%%%%%%%

%%%%%%%%%%%%%%%%%%%%%%%%%%%%%%%%%%%%%%%%%%%%%%%%%%%%%%%%%%%%%%%%%%%%%%%%%%%%%%%%%%%%%%%%%%%%%%%%%%%
\section{Meta Filter Pruning}

\subsection{Preliminaries}
First, we assume that a neural network has $L$ layers and the number of input and output channels in $i_{th}$ convolution layer is $N_{i}$ and $N_{i+1}$, respectively.
Suppose $K$ is the kernel size of the layer, we use $\mathcal{F}_{i,j}$ to represent the $j_{th}$ filter in the $i_{th}$ layer, and $\mathcal{F}_{i,j}\in \mathbb{R}^{N_{i}\times K\times K}$.
For the $i_{th}$ layer, it consists of a set of filters denoted by $\{\mathcal{F}_{i,j}, 1 \leq j \leq N_{i+1}\}$ and parameterized by  $\{\mathbf{W}^{(i)}\in\mathbb{R}^{N_{i+1}\times N_{i}\times K\times K},1\leq i\leq L\}$. 
% Assume the shapes of input tensor $\mathbf{I}$ and output tensor $\mathbf{O}$ are $N_{i} \times H_{i}\times W_{i} $ and $N_{i+1} \times H_{i+1}\times W_{i+1} $, respectively.
% The convolutional operation of the $i_{th}$ layer can be written as:
%\footnote{Fully-connected layers equal to convolutional layers with $k=1$}

% \vspace{-4mm}
% {\small
% \begin{align}
% \label{eq:1}
% \mathbf{O}_{i,j}=\mathcal{F}_{i,j}\ast\mathbf{I} ~&~ \mathbf{for}~1 \leq j \leq N_{i+1},
% \end{align}
% }

% \vspace{-2mm}

% \noindent
% where $\mathbf{I}$ is the input tensor with a shape of $N_{i} \times H_{i}\times W_{i} $, $\mathbf{O}$ is the output tensor with a shape of  $N_{i+1} \times H_{i+1}\times W_{i+1} $.

For the convenience of our discussion, we assume $\mathcal{F}_{i,j}$ consists of two subsets: the pruned filter set $\gF^{pruned}$, and the remaining filter set $\gF^{keep}$. Specifically:

\RRRR{
{\small
\begin{equation}
\label{eq:pruning3}
\begin{aligned}
\gF^{keep}
&= \big \{ \gF_{i,j}, \quad for \enspace  i \in [1,L],  \enspace j \in ID(i)  \enspace \big \} \\
\end{aligned}
\end{equation}
}
}

\vspace{-4mm}

\noindent
\RRRR{
where $ID(i)$ is the index of important filters in layer $i$. 
}

\vspace{2mm}

Given a dataset $\gD = \{(\rvx_i, \rvy_i)\}_{i=1}^n$, and a desired sparsity level $\kappa$ (\ie, the number of remaining non-zero filters), we solve a constrained optimization problem, defined as follows:

% then we have:

% {\small
% \begin{equation}
% \label{eq:filter-set}
% \begin{aligned}
% % &\gF^{keep} \cup \gF^{pruned}=\gF^{original}\\
% &\gF^{keep} \cup \gF^{pruned}=\gF \\
% &\gF^{keep} \cap \gF^{pruned}=\emptyset
% \end{aligned}
% \end{equation}
% }

% Assume the filter pruning rate for the $i_{th}$ weighted layer is $P_{i}$, $\gF^{keep}$ set is constructed by $N_{i+1}\times P_{i}$ filters which selected by a pruning criterion.
% The index of these selected filters are ${\sI}$, then 
% $\gF^{keep}    \equiv \mathcal{F}_{i,\sI}, \sI \in [1, N_{i+1}]$

% {\small
% \begin{equation}
% \label{eq:filter-select}
% \begin{aligned}
% &\gF^{keep} = u(\gF^{original},\|\mathcal{F}_{i,j}\|_{p},AveD({\pmb x}))\\
% \end{aligned}
% \end{equation}
% }

% which means the remaining filters is a function of $\gF^{original}$ and specicific pruning criterion such as $\|\mathcal{F}_{i,j}\|_{p}$ and $AveD({\pmb x})$.

%The filter ranking in the $i_{th}$ layer could be easily obtained from the above Equation~\ref{eq:p-norm-select} and Equation~\ref{eq:filter-select}.
%makes $\|\mathcal{F}_{i,j}\|_{p}$ and $AveD({\pmb x})$ to the smallest extend to construct the pruning set $\gF^{pruned}$.

\RRRR{
\vspace{-4mm}
\begin{equation}
\begin{aligned}\label{eq:nn}
\min_{\gF^{keep}} \ell (\gF^{keep};\gD) &= \min_{\gF^{keep}} \frac{1}{n} \sum_{i=1}^n \ell(\gF^{keep}; (\rvx_i, \rvy_i)) \\
\text{s.t.}  \quad N_{set} (\gF^{keep})  &\le \kappa ,  \quad \gF \in \R^{N \times K\times K}.
\end{aligned}
\end{equation}
}

\noindent
where $\ell(\cdot)$ is a standard loss function (\eg, cross-entropy loss), $\gF^{keep}$ is the set of remaining filters of the neural network, and $N_{set}$ is the cardinality of the filter set. 

% Obviously, $\gF^{keep}$ in Equation~\ref{eq:nn} can be easily determined if we have a corresponding $\gF^{pruned}$.

% In this section, we introduce the framework of meta filter pruning (MFP) as well as two new measurements for Relational criterion, so that we could choose a suitable filter pruning criterion at a specific training or re-training stage. 
%For simplicity and without loss of generality, we consider filters within a single layer in this section.

\subsection{Information From the Filters}\label{sec:Information}

For filter pruning, the most critical part is to evaluate the importance of the filters.
\RR{For filters with specific weight values, we could obtain the magnitude information and the geometric information of the filters.}

\subsubsection{Magnitude Information}\label{sec:Magnitude}
 Magnitude-based criterion believes that the convolutional results of the filter with the smaller $\ell_{p}$-norm lead to relatively lower activation values~\cite{li2016pruning}, so the contribution of the small-norm filters to the final prediction of deep CNN models is less than that of large-norm filters. The $\ell_{p}$-norm of filters is:

{\small
\begin{align}\label{eq:p-norm2}
\|\mathcal{F}_{i,j}\|_{p}=\sqrt[p]{\sum_{n=1}^{N_{i}}\sum_{k_{1}=1}^{K}\sum_{k_{2}=1}^{K}\left|\mathcal{F}_{i,j} (n,k_{1},k_{2})\right|^{p}}
\end{align}
}

In terms of this understanding, we give a high priority to prune the filters of a small $\ell_{p}$-norm than filters of a higher $\ell_{p}$-norm, which could be formulated as:

\vspace{-2mm}
{\small
\begin{align}\label{eq:p-norm}
 j^* = \argmin_{j\in[1,N_{i+1}]} \|\mathcal{F}_{i,j}\|_{p},
\end{align}
}

\vspace{-2mm}

\subsubsection{\RR{Geometric Information}}\label{Sec:Correlation}

\RR{As the $\ell_{p}$-norm only models the magnitude information of the filters, we introduce the distance of filters to reflect the geometric information between them.}
We utilize the Minkowski distance~\cite{singh2013k} as our selection. For simplicity, we reshape or extend the three-dimensional filter $\mathcal{F}_{i,j}$ to one-dimensional vector. 
Then, the $i_{th}$ convolution layer could be written as $\gY \in \mathbb{R}^{N_{i+1}\times G_{i} }$, which means $N_{i+1}$ vectors and the length of each vector is $G_{i} =N_{i}\times K \times K$. If we choose two vectors $\pmb x, \pmb y \in \mathbb{R}^{1\times G_{i}}$.
Then the Minkowski distance between $\pmb x$ and $\pmb y$ is:

{\small
\begin{align}
\label{eq:Minkowski}
D(\pmb x, \pmb y) = \sqrt[q]{\sum_{i=1}^{G_{i}}\left|  x_{i}- y_{i}\right|^{q}},
\end{align}
}

\noindent
where $x_{i}$ is the $i_{th}$ element of the vector $\pmb x$.

\RRRR{Note that Minkowski distance is a general distance which can be considered as a generalization of both the Manhattan distance~\cite{sinwar2014study} and the Euclidean distance~\cite{gower1985properties}.
Therefore, it is easy to cover different kinds of distance by adjusting the value of the parameter $q$ in Eq.~\ref{eq:Minkowski}.
That is to say, it is simple for us to increase the candidate filter importance measurement methods, thus enlarging our search space.
For example, if we use $q=1$ in Eq.~\ref{eq:Minkowski}, we get the Manhattan distance.
If we use $q=2$ in Eq.~\ref{eq:Minkowski}, we obtain the Euclidean distance.
}

For simplicity, we discuss pruning only one filter of a network layer with $N$ filters.
To reduce the accuracy gap between the pruned model and the original model, the remaining $N-1$ filter should best approximate the original $N$ filters.
In other words, the contribution of this pruned filter should be replaced by the remaining $N-1$ filters.
\RR{Based on this consideration, we prefer to remove the filter that minimizes the average distance to other filters.}
% , as the filters contain redundant information and can be replaced by other filters.}
To this end, we define the average distance for a filter as the importance evaluation, which could be formulated as:

\RRRR{
{\small
\begin{equation}\label{eq:aved}
D_{avg}({\gY_{p}}) = \frac {\sum_{p=1,p \neq q}^{N_{i+1}} D(\gY_{p}, \gY_{q})}{N_{i+1}} = \frac {\sum_{p=1}^{N_{i+1}} D(\gY_{p}, \gY_{q})}{N_{i+1}},
\end{equation}
}
}

\noindent
\RRRR{where $\gY_{p}$ is the $p_{th}$ filter vector of the layer. So the index of the pruned filter could be formulated as:}

\RRRR{
{\small
\begin{equation}
\label{eq:filter-select}
\begin{aligned}
p^* =  \argmin_{p\in[1,N_{i+1}]}D_{avg}({\gY_{p}}).\\
\end{aligned}
\end{equation}
}
}
%-------------------------------------------------------

%-------------------------------------------------------

% \subsubsection{Datasets and Meta-attributes of Pruning}\label{Meta-attributes}
\subsection{Meta-attribute based Pruning}\label{sec:framework}

%------------------------------------------------------------

\setlength{\tabcolsep}{0.85em}
\begin{table*}[!t]\small
\centering  	
\begin{tabular}{|c|c c c c c c c|}  		
\hline 		
Depth & Method &Pre-train? &Baseline acc. (\%)  &Accelerated acc. (\%)  &Acc. $\downarrow$ (\%) & FLOPs & FLOPs $\downarrow$(\%)       
\\ 
%\hline \hline   

%\multirow{2}{*}{20}        
%&MIL~\cite{Dong_2017_CVPR} &\xmark        & 91.53        & \textbf{91.43} &0.10 &	3.20E7	  & 20.3	 \\          
%& SFP~\cite{he2018soft} & \xmark       & \textbf{92.20} ($\pm$0.18)       &  90.83  ($\pm$0.31)& 1.37&  {2.43E7} &{42.2}  \\
% & Ours (MFP-only 40\%) & \xmark       & \textbf{92.20} ($\pm$0.18)       &  {90.34} ($\pm$0.16) & {1.86}&  \textbf{1.87E7} &\textbf{54.0}  \\  

\hline     \hline         

\multirow{2}{*}{32} &MIL~\cite{Dong_2017_CVPR} &\xmark       &92.33         & 90.74 &1.59 &	4.70E7	  & 31.2	 \\          
%&SFP~\cite{he2018soft}  & \xmark    & \textbf{92.63} ($\pm$0.70)       &  92.08 ($\pm$0.08)& 0.55 &  {4.03E7} &{41.5} \\ 
& Ours (40\%)   & \xmark    & \RR{92.63} ($\pm$0.70)       &  \textbf{91.85} ($\pm$0.09) & \textbf{0.78}&  \textbf{3.23E7} &\textbf{53.2} \\

\hline  \hline

\multirow{8}{*}{56} 		
&PFEC~\cite{li2016pruning} &\xmark      & 93.04                   & 91.31  &1.75 &9.09E7  &27.6		 \\         
&CP~\cite{He_2017_ICCV} &\xmark       & 92.80                      & 90.90 &1.90 &	\textbf{--}	  & 50.0	 \\         

&  SFP~\cite{he2018soft}  &\xmark  	&\RR{93.59} ($\pm$0.58)  & 92.26  ($\pm$0.31)& 1.33   &  \textbf{5.94E7} &\textbf{52.6} \\

& Ours (40\%)    &\xmark  	&\RR{93.59} ($\pm$0.58)  & \textbf{92.76} ($\pm$0.03) & \textbf{0.83}   &  \textbf{5.94E7} &\textbf{52.6} \\   

\cdashline{2-8} 

&PFEC~\cite{li2016pruning} &\cmark      & 93.04          & 93.06 &\textbf{-0.02} &9.09E7 & 27.6 \\	
& \RRR{ DSA~\cite{ning2020dsa}}  & \RRR{\cmark}  	&\RRR{93.12}  & \RRR{92.91}  &\RRR{0.22}  &  \RRR{\textbf{--}} & \RRR{\textbf{49.7}} \\       

&CP~\cite{He_2017_ICCV} &\cmark        & 92.80                   & 91.80 &1.00 &	\textbf{--}	 &	50.0 	 \\         
&AMC~\cite{he2018amc} &\cmark        & 92.80                   & 91.90 &0.90 &	\textbf{--}	 &	50.0 	 \\         
& \RRR{ NPPM~\cite{gao2021network}}  & \RRR{\cmark}  	&\RRR{93.59 ($\pm$0.58)}  & \RRR{93.40 ($\pm$0.34)}  &\RRR{0.19}  &  \RRR{\textbf{--}} & \RRR{\textbf{50.0}} \\       
& \RRR{ SCP~\cite{kang2020operation}}  & \RRR{\cmark}  	&\RRR{93.59 ($\pm$0.58)}  & \RRR{93.23}  &\RRR{0.36}  &  \RRR{\textbf{--}} & \RRR{\textbf{51.5}} \\

& FPGM~\cite{he2018pruning}   &\cmark  	&\RR{93.59} ($\pm$0.58)   &{93.26} ($\pm$0.03) &{0.33}   & \textbf{5.94E7} &\textbf{52.6} \\ 

& \RRR{ CLR~\cite{le2021network}}  & \RRR{\cmark}  	&\RRR{93.59 ($\pm$0.58)}  & \RRR{92.78 ($\pm$0.34)}  &\RRR{0.81}  &  \RRR{\textbf{5.94E7}} & \RRR{\textbf{52.6}} \\

& Ours (40\%)   &\cmark  	&\RR{93.59} ($\pm$0.58)   &\textbf{93.56} ($\pm$0.16) &{0.03}   & \textbf{5.94E7} &\textbf{52.6} \\      

\hline     \hline          

\multirow{9}{*}{110}          		
&PFEC~\cite{li2016pruning} &\xmark   &93.53     & 92.94      &  0.61              & 1.55E8 	&38.6 	 \\          
&MIL~\cite{Dong_2017_CVPR} &\xmark        & 93.63         & {93.44} &{0.19} &	-	  & 34.2 	 \\          
& SFP~\cite{he2018soft}    &\xmark  & \RR{93.68} ($\pm$0.32) 	& 93.38  ($\pm$0.30)& 0.30& {1.50E8} &{40.8}   \\  		
& Rethink~\cite{liu2018rethinking}  &\xmark  & \RR{93.77} ($\pm$0.23) 	& \textbf{93.70}  ($\pm$0.16)& {0.07}  & {1.50E8} &{40.8}   \\  

& Ours (40\%)   &\xmark  & \RR{93.68} ($\pm$0.32) 	&  {93.69} ($\pm$0.31) &  \textbf{-0.01} & {1.21E8} &{52.3}\\ 
& Ours (50\%)   &\xmark  & \RR{93.68} ($\pm$0.32) 	&  {93.38} ($\pm$0.16) &  {0.30} & \textbf{9.40E7} &\textbf{62.8}\\ 

\cdashline{2-8} 

&PFEC~\cite{li2016pruning} &\cmark     & 93.53    & 93.30  &{0.20} 	&1.55E8 	&38.6\\ 
&NISP~\cite{yu2018nisp} &\cmark        & \textbf{--}                   & \textbf{--} &\textbf{0.18} &	\textbf{--}	 &	43.8 	 \\         
&\RRR{CLR~\cite{le2021network}} & \RRR{\cmark}       & \RRR{93.68 ($\pm$0.32)} 	& \RRR{92.91 ($\pm$0.41)} &  \RRR{0.77} &	 \RRR{\textbf{1.21E8}}	 & 	 \RRR{52.3} 	 \\         

& Ours (40\%)  &\cmark  & \RR{93.68} ($\pm$0.32) 	& \textbf{93.31} ($\pm$0.08) &{0.37}& \textbf{1.21E8} &\textbf{52.3}   \\  		

\hline  	
\end{tabular}  	
\caption{Comparison of the pruned ResNet on CIFAR-10.
In ``Pre-train?'' column, ``\cmark'' and ``\xmark'' indicate pruning the pre-trained and scratch model, respectively.
The ``Acc.~$\downarrow$'' is the accuracy drop between the pruned model.
%A negative value in ``Acc.~$\downarrow$'' indicates an improved model accuracy.
} 	
\label{table:cifar10_accuracy} 
\end{table*}  

Existing works~\cite{li2016pruning,he2018soft} prune the filters based on a pre-defined criterion, without considering the change of filter distribution due to network weights updating.
In this section, we illustrate our proposed meta-attribute based pruning method, which can adaptively choose an appropriate criterion based on the current state of the network.

% \textbf{Traditional Meta-learning} 
% Given a query dataset, a standard meta-learning method selects a similar dataset from previously processed datasets (training datasets). Then the performance of a candidate algorithm of the query dataset and the chosen training dataset should be similar~\cite{brazdil2003ranking}.
% Therefore, measuring the similarity between datasets is a crucial part of traditional meta-learning.

\RRRR{As shown in \cite{brazdil2003ranking}, the general, statistical and information theoretic \emph{measures} to characterize the datasets are called \emph{meta-attributes} of the datasets.}
In the filter pruning scenario, we should minimize the difference of the meta-attributes in the original network ($\gF$) and those in the pruned network ($\gF^{keep}$).
If the meta-attributes of these two networks are similar, $\gF^{keep}$ could achieve performance as good as $\gF$ do.

%Note that there exists a difference between traditional meta-learning and our proposed meta-pruning. The former setting has multiple candidates, while our setting only has one for comparison, \ie, the original model $\gF$.

Previous works provided by~\cite{He_2017_ICCV,he2018soft} combine the pruning process and training process, then conduct pruning in an iterative manner.
We follow them and model the meta-process of filter pruning as a sequential decision process in a greedy approach. The entire pruning and training process is illustrated in~\Figref{fig:process}, which is elaborated as follows.
\begin{itemize}

\item Assume $S$ is a set of states. In our case, $s_t\in S$ is the network at the $t_{th}$ training epochs. \RRRR{At each time step $t$, $s_t$ is available to the meta-process which decides the remaining filter set $\gF^{keep}_{t}$.}

\RRRR{
\item At the $t$-th step, given the state $s_t$, the meta-process takes an action to choose the proper pruning criteria for filter importance evaluation. 
We use $\mathcal{A}_t$ to indicate the action of the $t$-th step.
Assume we have $C$ candidate pruning criteria, then $ \mathcal{A}_t \in \mathbb{R} ^ {C \times 1}$.
For each dimension of $\mathcal{A}_t$, we use the binary value {0,1} to indicate whether a specific pruning criterion is selected (1 means selected, and 0 means not selected). In our setting, only one pruning criterion is chosen for the current state, which is: }

\vspace{-2mm}
\RRRR{
{\small
\begin{equation}
\label{eq:one-criterion}
\begin{aligned}
\sum_{i=1}^{C} \mathcal{A}_{t,i} =1,\quad \mathcal{A}_{t,i}\in \{0,1\},
\end{aligned}
\end{equation}
}
}
\vspace{-2mm}

\RRRR{
where $\mathcal{A}_{t,i}$ means the indicator of the $i_{th}$ criteria in the $t_{th}$ step.
}

\RRRR{
\item $\phi: S\rightarrow \mathcal{A}$ is the policy employed by the meta-process to generate its action:  $\phi(s_t)=\mathcal{A}_t$. For filter pruning, the policy should aim at reducing the difference between meta-attributes of the pruned models and the original models:
}

% In the filter pruning scenario, it is the network that should be measured.
% Therefore, we treat the original model ($\gF^{original}$) and the pruned model ($\gF^{keep}$) as datasets.
% If the meta-attributes of these two models are similar, then the corresponding performances of them are similar. In this way, $\gF^{keep}$ could achieve good performance on the proposed task as $\gF^{original}$ do.
% Note that there exists difference between traditional meta-learning and meta-pruning. The former setting has multiple candidate dataset, while the latter setting only has one dataset for comparison, \ie, the original model $\gF^{original}$.
% Now we should maxmize the similarity of $\gF^{original}$ and $\gF^{keep}$,  \ie, minimize the meta-attributes of these models:

\RRRR{
{\small
\begin{equation}
\label{eq:min-attribute}
\begin{aligned}
& \mathcal{A}_t^{*} = \argmin_{\mathcal{A}_t}{ \mid \rmM (\mathcal{F}_{ \mathcal{A}_t } ) - \rmM (\mathcal{F}) \mid} ,
\end{aligned}
\end{equation}
}
}

\vspace{-4mm}

\RRRR{
\noindent
where $\rmM$ represents the meta-attributes of the network, $\mathcal{F}_{ \mathcal{A}_t}$ is the pruned network under a specific criterion from the action $\mathcal{A}_t$.
Several measures could be utilized as meta-attributes, such as sparsity level $\kappa$, the mean value of weights, top-5 loss, top-1 loss, and so on.
}

%From the empirical analysis, we find the top-5 loss is the best meta-attributes, as shown in Section~\ref{Sec:Case}.

\item After we enter time step $t+1$, we follow the above protocol and take action $a_{t+1}$ based on the state $s_{t+1}$. 
\end{itemize}

\begin{algorithm}[t]
\caption{Algorithm Description of MFP}
\label{alg:MFP}

\begin{algorithmic}[1]
\INPUT training data: $\mathbf{X}$, model $\mathbf{W} = \{\mathbf{W} ^{(i)}, 0\leq i \leq L\}$.
\State  \textbf{Given}:  pruning rate $P_{i}$
\State  \textbf{Initialize}: model parameter $\mathbf{W}$
\For{$epoch=1$; $epoch \leq epoch_{max}$; $epoch++$}
	\State Update the model parameter $\mathbf{W}$ based on $\mathbf{X}$
	\State Find $\mathcal{A}^{*}$ that satisfy Eq.~\ref{eq:min-attribute}

	\For{$i=1$; $i \leq L $; $i++$}
		\State Using $\mathcal{A}^{*}$  to prune $N_{i+1}P_i$  filters
	\EndFor
\EndFor
\State Obtain the compact model $\mathbf{W} ^{*}$ from $\mathbf{W}$
\OUTPUT The compact model and its parameters $\mathbf{W} ^{*}$
\end{algorithmic} 
\end{algorithm}

The algorithm of meta filter pruning is shown in Algorithm~\ref{alg:MFP}. 
\RR{Although there are a few similarities between our algorithm and evolving Takagi–Sugeno (TS) fuzzy model~\cite{kalhor2013evolving}, there are still significant differences between them.
Specifically, a TS model defines neighboring models as a certain number of models with split and merge operations. Then the TS model switches to one of the neighboring models at each stage of evolving. 
In contrast, our method treats models with different meta-attributes as neighboring models, and we select one of those models during our training process.}
We zeroize the selected filters to maintain the model capacity of the network~\cite{he2018soft}, but the process has the same effect as pruning~\cite{li2016pruning}.

\subsection{\RR{Scope of Search}}

\subsubsection{\RR{Reasonable Filter Ranking Criteria}}

\RR{
If we define the search scope by directly selecting a number of filters, the search scope is surprisingly large for those frequently used CNN architectures. For example, if we want to keep $n$ filters in $i_{th}$ layer, which has a total of $N_{i+1}$ filters, then the number of selections would be $\dbinom{N_{i+1}}{n} = \frac{N_{i+1}!} {n! \enspace (N_{i+1}-n)!}$, where $\dbinom {}{}$ denotes the combination~\cite{benjamin2003proofs}. 
We take ResNet-50 for example. The first layer of ResNet-50 has 64 filters and the total number of selecting 10 filters is $\dbinom {64}{10}=151,473,214,816$ selections. The upper layers with more filters would have a much higher selection number. As a result, the number of the candidate neighboring CNNs is very large and the computational cost is not affordable if we conduct brute-force searching.
 }

\RR{
Considering the above, adapting the filter ranking criteria is necessary to reduce the searching space. 
In this way, the scope of search is mainly decided by the number of reasonable filter ranking criteria.
Suppose we have $S$ filter ranking criteria, the number of selecting $n$ filters from $N_{i+1}$ filters would decrease from $\dbinom{N_{i+1}}{n}$ to $S$, which is a reasonable size of the search space.
}
 
\subsubsection{\RR{Clear Model Judgment}}
\RRRR{
Another factor we need to consider is that the model judgment should be conducted clearly.} Suppose the number of candidate acceleration ratios for pruned models is $R$. For every acceleration ratio, the number of neighboring CNNs is $S$ considering the number of criteria is $S$. So the total number of candidate CNNs is $R\times S$.

\RR{
We consider two aspects to judge the performance of a model, \ie, accuracy and acceleration ratio (FLOPs). It is difficult to evaluate since those two aspects are contradictory to each other. For example, suppose a model has a 10\% acceleration ratio with a 1\% accuracy drop, and another model has a 20\% acceleration ratio with a 2\% accuracy drop. In this situation, it may be difficult to choose between the two models. To solve this problem, we fix one of the two aspects, i.e., acceleration ratio, and utilize the accuracies under the same acceleration ratio to judge the model performance. In this way, the total number of candidate CNNs for a specific acceleration ratio is $S$.
% If the two aspects of the two models are different, it is difficult to judge since both aspects are always contradictory to each other. 
}

\subsection{Acceleration Analysis}\label{Calculation Reduction After Pruning}
In the above analysis, the ratio of pruned FLOPs is $1-(1-P_{i+1})\times(1-P_{i})$ theoretically. As other operations such as batch normalization (BN) and pooling are insignificant compared to convolution operations, it is common to utilize the FLOPs of convolution operations as the FLOPs of the network~\cite{li2016pruning,he2018soft}.

However, in the real scenario, non-tensor layers (\eg, pooling and BN layers) also need the computation time on GPU~\cite{Luo_2017_ICCV} and the realistic acceleration may be influenced.
Besides, other factors such as buffer switch, IO delay and the efficiency of BLAS libraries also lead to the gap between the realistic and theoretical acceleration. We compare the different acceleration ratios in Table~\ref{table:Comparison_Speed}.

%-------------------------------------------------------------------------
\section{Experiments}\label{Experiment}

%-------------------------------------------------------

\setlength{\tabcolsep}{0.65em} 
\begin{table*}[!h] \small 
\centering 	
\begin{tabular}{|c |c c c c c c c c c|} 		
\hline    		
\multirow{3}{*}{Depth}	   & \multirow{3}{*}{Method}  &\multirow{3}{*}{\shortstack {Pre-\\train?}}  &\multirow{3}{*}{\shortstack {Baseline\\top-1\\acc.(\%)} }  &\multirow{2}{*}{\shortstack {Accelerated\\top-1\\acc.(\%)} } &\multirow{2}{*}{\shortstack{Baseline\\top-5\\acc.(\%)} }   &\multirow{2}{*}{\shortstack {Accelerated\\top-5\\acc.(\%)} }   &\multirow{3}{*}{\shortstack {Top-1 \\acc. $\downarrow$(\%)} }  &\multirow{3}{*}{\shortstack {Top-5\\acc. $\downarrow$(\%)} } & \multirow{3}{*}{\shortstack {FLOPs$\downarrow$\\(\%)}}    \\
& & & & & & & & & \\
& & & & & & & & & \\
 \hline      \hline  
            
\multirow{5}{*}{18} &MIL~\cite{Dong_2017_CVPR} &\xmark   &69.98 &66.33 & 89.24     & 86.94	&3.65	 &2.30   & 34.6		 \\	       		  
%& Ours(30\%) &N &\textbf{70.23$\pm$0.06}	&\textbf{67.25$\pm$0.13} & \textbf{89.51$\pm$0.10}  & \textbf{87.76$\pm$0.06}   &\textbf{2.98}   & \textbf{1.75} & \textbf{41.8} \\\hline       
&SFP~\cite{he2018soft} &\xmark &\RR{70.28}	&{67.10} & \RR{89.63}  & {87.78}   &{3.18}   & {1.85} & \textbf{41.8} \\

& Ours (30\%) &\xmark &\RR{70.28}	&\textbf{67.66} & \RR{89.63}  & \textbf{87.90}   &\textbf{2.62}   & \textbf{1.73} & \textbf{41.8} \\
\cdashline{2-10}

&Ours (30\%) &\cmark &\RR{70.28}	&\textbf{68.31} & \RR{89.63}  & \textbf{88.28}   &\textbf{1.97}   & \textbf{1.35} & {41.8} \\
&\RRRR{COP~\cite{wang2019cop}} & \RRRR{\xmark} & \RRRR{{70.29}}	& \RRRR{{66.98}} & \RRRR{\textbf{--}}	  & \RRRR{\textbf{--}}	   &\RRRR{3.31}   & \RRRR{\textbf{--}}	 & \RRRR{43.3} \\

& Ours (40\%) &\cmark &\RR{70.28}	&{67.11} & \RR{89.63}  & {87.49}   &{3.17}   & {2.14} & \textbf{51.8} \\

     \hline          \hline     

 \iffalse
 
\multirow{6}{*}{34}	          
%&~\cite{Dong_2017_CVPR} &N   & 73.42          &\textbf{72.99} 	&91.36	 &\textbf{91.19}	&\textbf{0.43} 	&\textbf{0.17} 	& 24.8	\\  
&SFP~\cite{he2018soft}	&\xmark 	&\textbf{73.92}		&71.83	&\textbf{91.62}   & 90.33   & 2.09  & 1.29 & \textbf{41.1}    \\    
 
 &  Ours (MFP-only 30\%)	&\xmark	&\textbf{73.92}		&71.79	&\textbf{91.62}   &\textbf{90.70}   & {2.13}  &\textbf{0.92}   & \textbf{41.1}    \\     
   &  Ours (MFP-mix 30\%)	&\xmark	&\textbf{73.92}		& \textbf{72.11}	&\textbf{91.62}   & {90.69}  &\textbf{1.81}    &{0.93}    & \textbf{41.1}    \\ 
% &  \BB{Ours -new-pure-dist(40\%)}	&\xmark 	&\textbf{73.92}		&70.78	&\textbf{91.62}   & 89.96   & {3.14}  &{1.66}   & \textbf{52.7}    \\     
   \cdashline{2-10} 
 &PFEC~\cite{li2016pruning} &\cmark    & {73.23}          & 72.17 	&\textbf{--}	 &\textbf{--} &\textbf{1.06} 	&\textbf{--} 	& 24.2	 \\ 

 &  Ours (MFP-only 30\%)	&\cmark &\textbf{73.92}		& {72.54}	&\textbf{91.62}   & \textbf{91.13}  & {1.38}   &\textbf{0.49}   & \textbf{41.1}    \\    
 
 & Ours (MFP-mix 30\%)	&\cmark 	&\textbf{73.92}		&\textbf{72.63}	&\textbf{91.62}   & {91.08}  &{1.29}    & {0.54}  & \textbf{41.1}    \\    
% &  \BB{Ours -new-pure-dist(40\%)}	&\cmark 	&\textbf{73.92}		&71.59	&\textbf{91.62}   & 90.49   & {2.33}  &{1.13}   & \textbf{52.7}     \\     
 \hline      \hline     
\fi

\multirow{10}{*}{50}	
%&SFP~\cite{he2018soft} & \xmark &\textbf{76.15}		&{74.61}		&\textbf{92.87}	  & {92.06} &1.54	 & 0.81 & 41.8  	 \\
&   Ours (40\%) & \xmark  &\RR{76.15}		&{74.13}		&\RR{92.87}	  & {91.94} & {2.02}	 &{0.93}  & {53.5}  	 \\

\cdashline{2-10}

& \RRRR{Imp~\cite{molchanov2019importance}} & \RRRR{\cmark} 	&\RRRR{{76.18}}	&\RRRR{{74.48}} & \RRRR{\textbf{--}}   &\RRRR{\textbf{--}}  	&\RRRR{1.70}	&\RRRR{ \textbf{--}}	 &\RRRR{35.0}	 	\\

&ThiNet~\cite{Luo_2017_ICCV}  &\cmark  &72.88 	&72.04  & 91.14   & 90.67              & {0.84} 	&{0.47}	& 36.7  \\  		  
&SFP~\cite{he2018soft} & \cmark  &\RR{76.15}		&{62.14}		&\RR{92.87}	  & {84.60} &14.01	 & 8.27 & 41.8  	 \\
%& FPGM~\cite{he2018pruning}       &\cmark &\textbf{76.15}		&{75.50}		&\textbf{92.87}	  & {92.63} &{0.65} 	 &{0.21}  & 42.2  	 \\
&\RRR{CLR~\cite{le2021network}} &\RRR{\cmark} 	&\RRR{76.15}	&\RRR{75.26} & \RRR{\textbf{--}}   &\RRR{\textbf{--}}  	&\RRR{0.89}	& \RRR{\textbf{--}}	 &\RRR{43.0}	 	\\     		  
&NISP~\cite{yu2018nisp} & \cmark  &\textbf{--}		&\textbf{--}	&\textbf{--}	  & \textbf{--} & 0.89	 & \textbf{--}& 44.0  	 \\

&CP~\cite{He_2017_ICCV} &\cmark 	&\textbf{--}	&\textbf{--} & 92.20    &90.80  	&\textbf{--}	& 1.40	 &{50.0}	 	\\     		  
&LFC~\cite{singh2018leveraging} &\cmark 	&75.30	&73.40  &92.20     	&91.40	& 1.90	& 0.80	 &{50.0}	 	\\     		
&ELR~\cite{wang2018exploring} &\cmark 	&\textbf{--}	&\textbf{--} & 92.20    &91.20  	&\textbf{--}	& 1.00	 &{50.0}	 	\\     		  
&\RRR{DSA~\cite{ning2020dsa}} &\RRR{\cmark} 	&\RRR{\textbf{--}}	&\RRR{\textbf{--}} & \RRR{\textbf{--}}   &\RRR{\textbf{--}}  	&\RRR{1.33}	& \RRR{\textbf{--}}	 &\RRR{50.0}	 	\\     		  

&\RR{Meta~\cite{liu2019metapruning}} &\cmark 	&{76.60}	&{75.40} & \textbf{--}   &\textbf{--}  	&{1.20}	& \textbf{--}	 &{51.2}	 	\\

&  Ours (30\%)  & \cmark  &\RR{76.15}		&\textbf{75.67}		&\RR{92.87}	  & \textbf{92.81} & \textbf{0.48}	 &\textbf{0.06}  &{42.2}   	 \\

&  Ours (40\%)  & \cmark  &\RR{76.15}		&{74.86}		&\RR{92.87}	  & {92.43} & {1.29}	 &{0.44}  &\textbf{53.5}   	 \\

\hline   
\iffalse
\multirow{2}{*}{101}

%& Ours (30\%) & \xmark  &\textbf{77.37}		& 77.03 	&\textbf{93.56}	  &  93.46  & 0.34 	 &  0.10  & {42.2}    \\ 
%\cdashline{2-10}ijump2(){}

& Rethinking~\cite{ye2018rethinking}  & \cmark  &\textbf{77.37}		& 75.27 	&\textbf{--}  &  \textbf{--}  & 2.10 	 &  \textbf{--}  & \textbf{47.0}    \\  
& Ours (MFP-only 30\%) &\cmark  &\textbf{77.37}		&\textbf{77.32}		&\textbf{93.56}	  & \textbf{93.56} &\textbf{0.05}	 & \textbf{0.00}& {42.2}    \\ 	
\hline    	
\fi

\end{tabular} 	
\caption{
Comparison of the pruned ResNet on ImageNet.
``Pre-train?'' and "acc. $\downarrow$" have the same meaning with Table~\ref{table:cifar10_accuracy}.
%We run ResNet-18 for three times to get the mean and std.
%For the other depths of ResNet, we just list the one-view accuracy.
} 
\label{table:imagenet_accuracy}
\end{table*}
%%%%%%%%%%%%%%%%%%%%%%%%%%%%%%%%%%%%%%%%%%%%%%%%%%%%%%%%%%%%%%%%%%%%%%%%%%%%%%%%%%%%%%%%%%%%%%%%
%-------------------------------------------------------

\subsection{Dataset and Architecture}

The experiments are conducted on two benchmark datasets, CIFAR-10~\cite{krizhevsky2009learning} and ILSVRC-2012~\cite{russakovsky2015imagenet}.
The CIFAR-10 dataset contains a total of $60,000$ images, which contains $50,000$ training images and $10,000$ testing images in $10$ different classes.
ILSVRC-2012~\cite{russakovsky2015imagenet} contains 1.28 million training images and 50k validation images of $1,000$ classes.

%As the ResNet has the shortcut structure, existing works~\cite{Dong_2017_CVPR,Luo_2017_ICCV,He_2017_ICCV} claim that ResNet has less redundancy than VGGNet~\cite{simonyan2014very} and accelerating ResNet is more difficult than accelerating VGGNet.
We focus on pruning the challenging multi-branch ResNet model~\cite{Dong_2017_CVPR,Luo_2017_ICCV,He_2017_ICCV}.
Moreover, to validate our method on the single-branch network, we follow~\cite{li2016pruning} to conduct a test on the VGGNet~\cite{cifar10_vgg}.

\subsection{Training and Pruning}\label{sec:exp-pruning}

\textbf{Training Setting.} For VGGNet on CIFAR-10, we follow the setting in~\cite{li2016pruning}. As the training setup is not publicly available, we re-implement the pruning procedure and achieve similar results to the original paper. For ResNet on CIFAR-10, we utilize the same training schedule as~\cite{zagoruyko2016wide}. For CIFAR-10 experiments, we run each setting three times and report the ``mean $\pm$ std''. In the ILSVRC-2012 experiments, we use the default parameter setting, which is the same as~\cite{he2016deep,he2016identity}, and the same data argumentation strategies as the official PyTorch~\cite{paszke2017automatic} examples.

\textbf{Pruning Setting.} For VGGNet on CIFAR-10, we use the same pruning rate as~\cite{li2016pruning}.
For experiments on ResNet, we follow~\cite{he2018pruning,he2018soft} and prune \emph{all} the weighted layers with the \emph{same} pruning rate at the same time.
Therefore, only one hyper-parameter, the pruning rate of $P_{i}=P$ is used to balance the acceleration and accuracy.
Note that choosing different rates for different layers could improve the performance~\cite{li2016pruning}, but it also introduces extra hyper-parameters.
Our pruning operation is conducted at the end of every two training epochs, which provides a balance of accuracy and energy consumption during the pruning operation.
Pruning of both the scratch model and the pre-trained model are compared.
For pruning from scratch, we train the model from scratch and do not conduct additional fine-tuning. For pruning the pre-trained model, the learning rate is reduced to one-tenth of the initial  learning rate.
The training schedule is the same as~\cite{he2018soft}.

\textbf{Meta-Process Setting.} \RR{
The execution time would increase when the number of neighboring CNN increases.
Specifically, the total execution time contains two parts. First, the time for the network training process. Second, the time for search at the end of the training epoch. Increasing the number of neighboring CNNs would affect the second part of the execution time, but not on the first part. 
For example, on the Quadro RTX 6000 GPU, training CIFAR-10 for one epoch takes 27.7 seconds, while searching for $N$ pruned models takes about $N \times 6.9$ seconds.
To ensure the searching time does not exceed the training time too much, the value of $N$ is 4. So we follow the definition of the scope of search in Section III.D and choose to search four pruned models. These models are obtained from four pruning criteria, including $p= 1,2$ in \Eqref{eq:p-norm2} for magnitude information, which are $\ell_{1}$-norm criteria~\cite{li2016pruning} and $\ell_{2}$-norm criteria~\cite{he2018soft}, respectively, and $q= 1,2$. in \Eqref{eq:Minkowski} for geometric information, which are Manhattan distance and the Euclidean distance, respectively.}
% In our experiment, we utilize the magnitude information and the geometric information to get the pruned model.
%  For magnitude information, we use $p=1$ and $p=2$ in Eq.~\ref{eq:p-norm}, 
% For geometric information, we utilize $q=1$ and $q=2$ in Eq.~\ref{eq:Minkowski} to obtain the Manhattan distance and the Euclidean distance, respectively.
Note that our framework is compatible with any new pruning criterion.
Based on the result in Sec.~\ref{Sec:Case}, we consider the top-5 loss as the meta-attributes to evaluate the network at the current state. Besides, the sparsity level $\kappa$ is utilized as a meta-attribute to balance the accuracy and the computational cost.

In Table~\ref{table:cifar10_accuracy} and Table~\ref{table:imagenet_accuracy}, the method labeled ``Ours (40\%)'' means we prune 40\% filters of the layers during training when the pruned filters is recovered.
We compare our method with existing state-of-the-art acceleration algorithms, \eg, MIL~\cite{Dong_2017_CVPR}, PFEC~\cite{li2016pruning}, CP~\cite{He_2017_ICCV}, ThiNet~\cite{Luo_2017_ICCV}, SFP~\cite{he2018soft}, NISP~\cite{yu2018nisp}, FPGM~\cite{he2018pruning}, LFC~\cite{singh2018leveraging}, ELR~\cite{wang2018exploring}, AMC~\cite{he2018amc}, \RRR{ DSA~\cite{ning2020dsa}, NPPM~\cite{gao2021network}, SCP~\cite{kang2020operation}, CLR~\cite{le2021network}}. Experiments show that our MFP achieves the comparable performance with the state-of-the-art results.

\subsection{VGGNet on CIFAR-10}
The result of pruning scratch and pre-trained VGGNet is shown in Table~\ref{table:vgg}. Not surprisingly, MFP achieves better performance than~\cite{li2016pruning} in both settings.
With the pruning criterion selected by our method, we could achieve better accuracy than~\cite{li2016pruning} when pruning the random initialized VGGNet (93.54\% \vs 93.31\%).
In addition, the pruned model without fine-tuning has better performance than~\cite{li2016pruning} (84.80\% \vs 77.45\%). 
After fine-tuning $40$ epochs, our model achieves similar accuracy with~\cite{li2016pruning}. Notably, if more fine-tuning epochs ($160$) are used,~\cite{li2016pruning} achieve similar result with fine-tuning $40$ epochs (93.28\% \vs 93.22\%), while our method could attain a much better performance (93.76\% \vs 93.26\%).

%%%%%%%%%%%%%%%%%%%%%%%%%%%%%%%%%%%%%%%%%%%%%%%%%%%%%%%%%%%%%%%%%%%%%%%%%%%%%%%%%%%%%%%%%%%%%%%%
\begin{figure*}[ht]
\center
\subfigure[Initialization 1]{
\label{fig:meta1}
\includegraphics[width=0.8\linewidth]{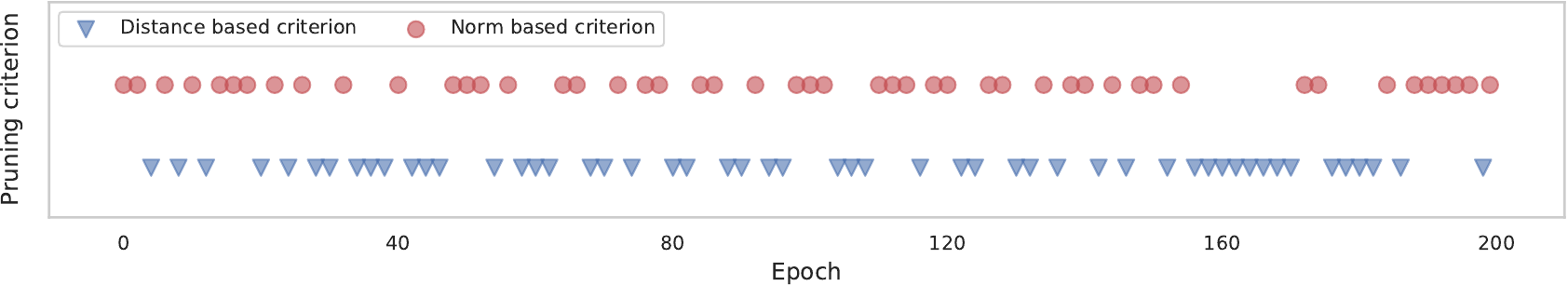}
}
\subfigure[Initialization 2]{
\label{fig:meta2}
\includegraphics[width=0.8\linewidth]{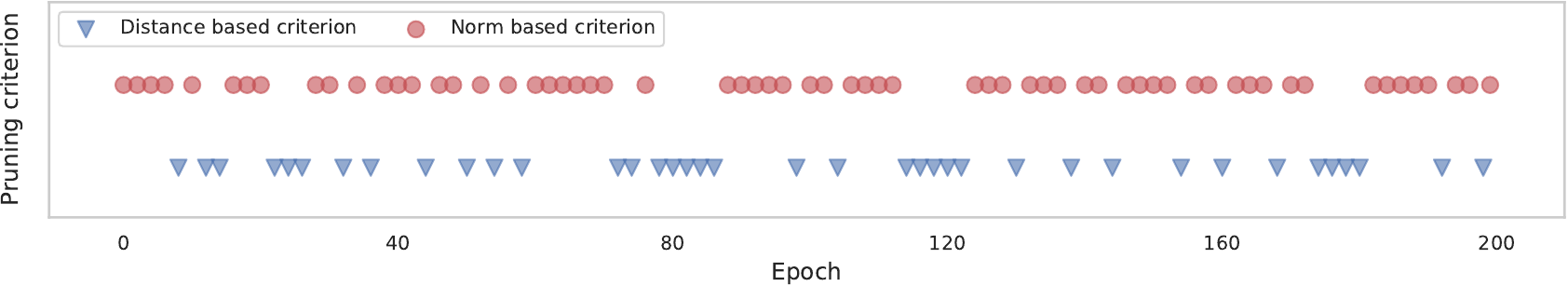}
}\caption{
Learned pruning criterion during the training process of ResNet-110 on CIFAR-10 under different initializations. The pruning rate is 40\%. The red and blue marker denotes the norm-based criterion and the distance-based criterion, respectively.
}\label{fig:meta}
\end{figure*}
%%%%%%%%%%%%%%%%%%%%%%%%%%%%%%%%%%%%%%%%%%%%%%%%%%%%%%%%%%%%%%%%%%%%%%%%%%%%%%%%%%%%%%%%%%%%%%%%

\begin{table}[ht]
\small
\setlength{\tabcolsep}{0.75em}
\begin{center}
\begin{tabular}{|c|c|c|}
\hline
{Setting $\backslash$ Acc (\%)} & PFEC~\cite{li2016pruning} & Ours  \\ \hline 
{Baseline }                   & 93.58 ($\pm$0.03)         & 93.58 ($\pm$0.03) \\\hline 
Prune from scratch                       & 93.31 ($\pm$0.03)         & \textbf{93.54} ($\pm$0.03) \\\hline 
Prune pretrain w.o. FT                & 77.45 ($\pm$0.03)         & \textbf{84.80} ($\pm$0.03) \\\hline
FT 40 epochs                  & 93.22 ($\pm$0.03)         & \textbf{93.26} ($\pm$0.03) \\\hline
FT 160 epochs                 & 93.28 ($\pm$0.03)         & \textbf{93.76} ($\pm$0.08) \\\hline 
\end{tabular}
\end{center}
\caption{Pruning scratch and pre-trained VGGNet on CIFAR-10. The abbreviation ``w.o. FT" means ``without fine-tuning".
}
\label{table:vgg}
\end{table}
%------------------------------------------------------

%------------------------------------------------------

\subsection{ResNet on CIFAR-10}\label{section:vgg}
%------------------------------------------------------

For the CIFAR-10 dataset, we test our MFP on ResNet (depth $32$, $56$ and $110$). We use two different pruning rates 40\% and 50\%. 
%All the convolutional layers are pruned with the same pruning rate at the same time.
As shown in Table~\ref{table:cifar10_accuracy}, the experiment results validate the effectiveness of our method.

\textbf{Result Explanation.} 
For example, comparing to SFP~\cite{he2018soft}, when we prune 52.6\% FLOPs of the scratch ResNet-56, our MFP has a 0.50\% accuracy improvement over SFP~\cite{he2018soft} (0.83\% \vs 1.33\%).
Besides, MIL~\cite{Dong_2017_CVPR} accelerates the scratch ResNet-32 by a 31.2\% speedup ratio with a 1.59\% accuracy drop, but our MFP achieves a higher speedup ratio with only less accuracy drop.
For pruning the pre-trained ResNet-56, our method achieves a higher acceleration ratio than CP~\cite{He_2017_ICCV} with a 0.97\% accuracy increase over CP~\cite{He_2017_ICCV}.
When compared with PFEC~\cite{li2016pruning}, our method achieves a higher speedup ratio even with accuracy improvement.
%accelerates the scratch ResNet-110 with a 52.3\% speedup ratio with even accuracy improvement, while PFEC~\cite{li2016pruning} achieves only 13.7\% less acceleration with 0.61\% accuracy drop. 
%This means PFEC~\cite{li2016pruning} harms the performance with a lower acceleration ratio.
%For NISP~\cite{yu2018nisp}, we achieve a higher speedup ratio (52.3\% \vs 43.8\%) with a similar accuracy drop.
\RRR{Comparing to DSA~\cite{ning2020dsa}, we achieve the 2.9\% more acceleration ratio with 0.19\% less accuracy drop on ResNet-56. When we achieve the same speed-up ratio with CLR~\cite{le2021network}, our accuracy is 0.74\% better than CLR~\cite{le2021network} on ResNet-56. Our performance is also better than NPPM~\cite{gao2021network} and SCP~\cite{kang2020operation} in terms of both accuracy and acceleration ratio.
}
Moreover, we achieve a better accuracy than~\cite{liu2018rethinking}, which is consistent with the conclusion by~\cite{liu2018rethinking} that different initializations would lead to different result networks.
\RR{The first reason for our superior result is that our criterion explicitly models the geometric information between filters by taking advantage of three corresponding measures.} Besides, we adaptively select the suitable criteria to match the current filter distribution, which may keep changing during the pruning process. Previous works~\cite{Dong_2017_CVPR,he2018soft,He_2017_ICCV,li2016pruning,yu2018nisp} pre-specify a criterion before pruning and keep it fixed during the entire pruning process, yet fail to consider that the selected criterion may no longer be suitable any more after epochs of training.

%%%%%%%%%%%%%%%%%%%%%%%%%%%%%%%%%%%%%%%%%%%%%%%%%%%%%%%%%%%%%%%%%%%%%%%%%%%%%%%%%%%%%%%%%%%%%%%%
\begin{figure}[!t]

\centering
\subfigure[Accuracy of ResNet-110 on CIFAR-10 with different FLOPs.]{
\label{fig:different_flop}
\includegraphics[width=0.7\linewidth]{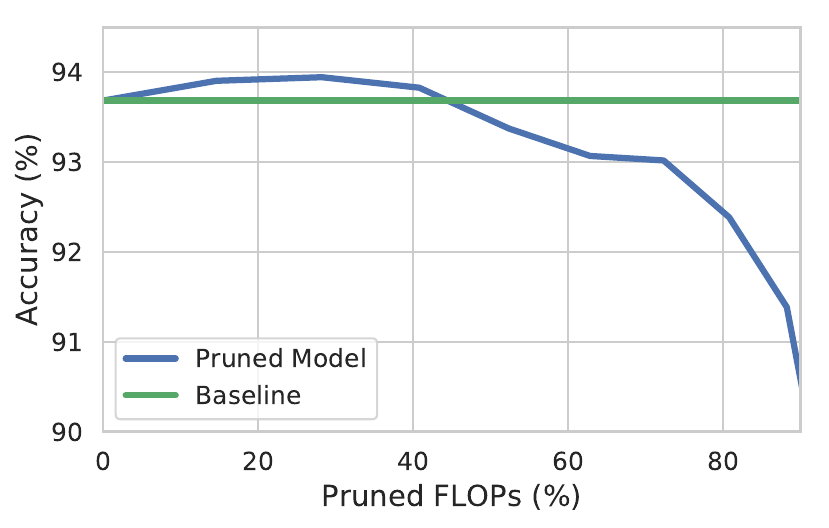}
}
\subfigure[Accuracy of ResNet-110 on CIFAR-10 with various pruning intervals.]{
\label{fig:different_epoch}
\includegraphics[width=0.7\linewidth]{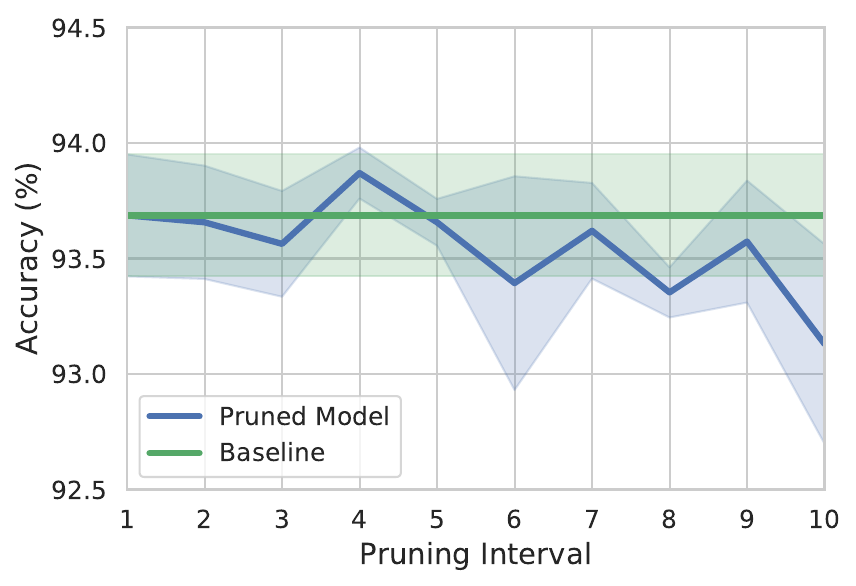}
}
\caption{Accuracy of ResNet-110 on CIFAR-10 regarding different FLOPs and pruning interval.}\label{Fig:abalation}
%\vspace{-2mm}
\label{fig:resnet_cifar100}
\end{figure}

\subsection{ResNet on ILSVRC-2012}
\label{section:ILSVRC}

For the ILSVRC-2012 dataset, we test our method on ResNet-18 and ResNet-50; and we use pruning rate 30\% and 40\% for these models.
Following~\cite{he2018soft}, we do not prune the projection shortcuts. 
%All the convolutional layer of ResNet are pruned with the same pruning rate at the same time.
The results are shown in Table~\ref{table:imagenet_accuracy}.

\textbf{Result Explanation.} 
For the random initialized ResNet-18, MIL~\cite{Dong_2017_CVPR} accelerates the network by a 34.6\% speedup ratio with a 3.65\% accuracy drop, but our MFP achieves a 41.8\% speedup ratio (7.20\% better) with only a 2.62\% accuracy drop (1.03\% better).
Comparing to SFP~\cite{he2018soft}, when we prune the same ratio (41.8\%) of FLOPs of the ResNet-18, our MFP has a 0.56\% accuracy improvement over SFP~\cite{he2018soft}.

For pruning the pre-trained ResNet-50, our MFP reduces 41.8\% FLOPs in the network with only a 0.06\% top-5 accuracy drop. In contrast, ThiNet~\cite{Luo_2017_ICCV} reduces 36.7\% FLOPs (5.1\% worse than ours) with a 0.47\% top-5 accuracy drop (0.41\% worse than ours).
In addition, SFP achieves the same acceleration ratio as MFP, with a 8.27\% top-5 accuracy drop (8.21\% worse than ours). 
Comparing to NISP~\cite{yu2018nisp}, we achieve a similar acceleration ratio with a smaller accuracy drop (0.48\% \vs~0.89\%).
When we prune 53.5\% FLOPs of the pre-trained ResNet-50, our MFP has 0.44\% top-5 accuracy drop, while CP~\cite{He_2017_ICCV} reduces 50.0\% FLOPs of the network with 1.40\% top-5 accuracy (0.96\% worse than ours).
\RRRR{
Comparing to Imp~\cite{molchanov2019importance}, we achieve a 18.5\% more acceleration ratio (53.5\% \vs~35.0\%) with a 0.41\% smaller accuracy drop (1.29\% \vs~1.70\%). When we achieve a similar FLOPs drop rate with COP~\cite{wang2019cop}, our accuracy drop is 1.34\% less than  COP~\cite{wang2019cop}. Also, We can obtain 8.5\% more acceleration ratio with 0.14\% less accuracy drop than COP~\cite{wang2019cop}.
}
\RR{We also achieve comparable results with MetaPruning~\cite{liu2019metapruning}, when our speed-up ratio is 2.3\% better than MetaPruning~\cite{liu2019metapruning} (53.5\% \vs~51.2\%), the accuracy drop is similar (1.29\% \vs~1.20\%).
The superior performance may come from that our method consider the magnitude information and the geometric information of the filters.}

%%%%%%%%%%%%%%%%%%%%%%%%%%%%%%%%%%%%%%%%%%%%%%%%%%%%%%%%%%%%%%%%%%%%%%%%%%%%%%%%%%%%%%%%%%%%%%%%%%

\subsection{Case Study}\label{Sec:Case}

%------------------------------------------------------
\begin{figure*}[ht]
\center
\includegraphics[width=0.99\linewidth]{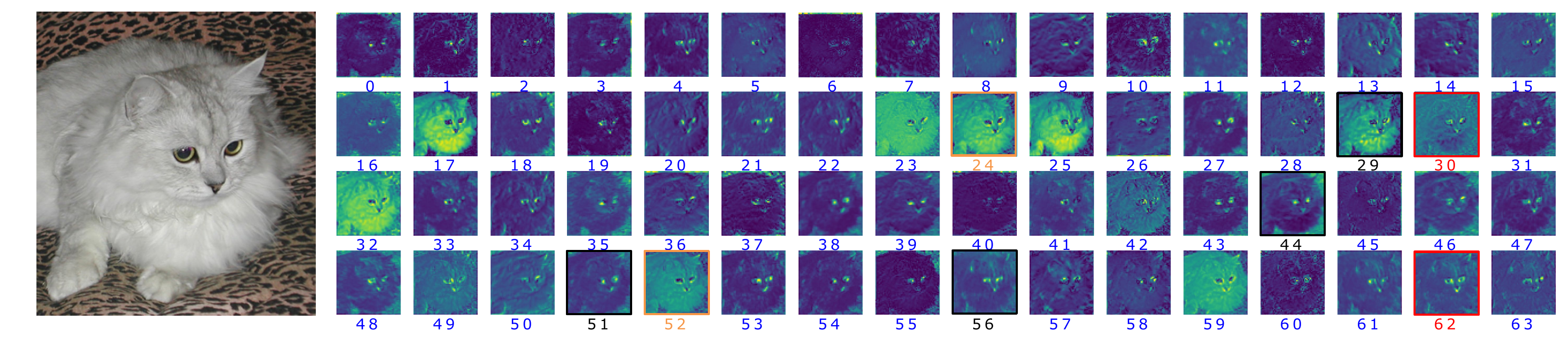}
\caption{
Input image (left) and visualization of feature maps (right, numbered 0 to 63).
Selected channels are pruned with norm-based criterion and relation-based criterion.
The feature maps are extracted from the first convolutional layer of the first block of ResNet-18. The pruning rate is 10\%.
Feature maps with a black title and box (channel 29, 44, 51, 56) denote the common channels selected by both criteria.
Feature maps with a red title and box (30, 62) denote the channels only selected by norm-based criterion, while feature maps with an orange title and box (24, 52) denote the channels only selected by distance-based criterion.
}
\label{fig:visual}
\vspace{-4mm}
\end{figure*}

\textbf{Different Criteria During Training.} The pruning criterion during the training process of two different initializations are shown in~\Figref{fig:meta}.
The norm-based criterion includes 1-norm and 2-norm. The relational criterion includes Manhattan distance and Euclidean distance. By comparing these two figures, we conclude that our MFP could adaptively select proper criteria during the training process with different initializations.
For the selected pruning criteria, we find that during the early training process, the distance-based criteria are adopted less than norm-based criteria. 
The above phenomena may be caused by the training knowledge from the training set.
\RR{During the early training stage, the filters have not learned enough training set knowledge, and the geometric information of filters is not particularly meaningful, so the norm-based criteria are preferred. After more training epochs, the filters obtain the information from the training set, and the geometric information of filters becomes meaningful; then the distance-based criteria begins to take effect.}

\textbf{Realistic Acceleration.} 
We measure the forward time of the pruned models to compare the theoretical and realistic acceleration. The experiment is conducted on one GTX1080 GPU with a batch size of 64 (Table~\ref{table:Comparison_Speed}).
As discussed in the above section, the gap between the theoretical and the realistic acceleration may come from the limitation of IO delay, buffer switch and efficiency of BLAS libraries.

\RRRR{
\textbf{Memory Footprint.} 
We test the memory footprint before and after pruning on NVIDIA RTX6000 GPU. For a batch size of 256, ResNet on CIFAR-10 requires 1423MB memory footprint before pruning. After pruning 52.3\% FLOPS, the required memory footprint is 840MB, which is 41.0\% less than the original memory footprint. The gap comes from operations such as batch normalization (BN) and pooling.
 }

\begin{table}[!t]
\small
\setlength{\tabcolsep}{0.4em}
\begin{center}
\begin{tabular}{| c | c | c | c  | c |}
\hline
\multirow{2}{*}{Model} & \multirow{2}{*}{\shortstack{Baseline\\time (ms)}} & \multirow{2}{*}{\shortstack {Pruned\\time (ms) }} & \multirow{2}{*}{\shortstack {Realistic \\Speedup(\%)}} &\multirow{2}{*}{\shortstack {Theoretical\\Speedup(\%)} }\\
               &         &          &        &          \\ \hline
ResNet-18      & 37.50   &  26.17   & 30.2   &  41.8    \\  
%ResNet-34      & 63.89   &  45.24   & 29.2   &  41.1    \\  
ResNet-50      & 136.24  &  84.33   & 38.1   &  53.5    \\ \hline
%ResNet-101     & 219.70  & 147.45   & 32.9   &  42.2    \\ 
\end{tabular}
\end{center}
\caption{
Comparison of the theoretical and realistic acceleration.
Only the time consumption of the forward procedure is considered.
%We use the GTX1080 GPU with a batch size of 64.
}
\label{table:Comparison_Speed}
\end{table}
\textbf{Varying Pruned FLOPs.}
We change the ratio of pruned FLOPs for ResNet-110 on CIFAR-10 to comprehensively understand our MFP, as shown in~\Figref{fig:different_flop}.
We are able to prune more than 40\% of the filters of the network without affecting the performance.
When the ratio of pruned FLOPs is less than 40\%, the performance of the pruned model even exceeds the baseline model without pruning. This means our MFP could choose the proper criterion and prune the suitable filters. In addition, our MFP may have a regularization effect on the neural network.

\textbf{Varying Pruning Interval.}
The pruning interval refers to the number of training epochs between two pruning operations.
We change the pruning interval from one epoch to ten epochs, as shown in Fig.~\ref{fig:different_epoch}, which illustrates the absence of large fluctuations in model accuracy across different pruning intervals.
This result means the performance of our framework is not sensitive to the pruning interval.

\textbf{Varying Meta-attributes.} We compare several meta-attributes to understand the MFP comprehensively.
The meta-attributes includes top5 loss, top1 loss, the mean value of the network, sparsity level $\kappa$, and so on.
\RRR{
Table \ref{table:mate-attr} shows the comparison between different meta-attributes.}
The sparsity level $\kappa$ meta-attributes is directly related to the acceleration ratio of the network. If we pre-define the expected acceleration ratio, the sparsity level $\kappa$ of the pruned model would be the same. Hence, we should consider other meta-attributes to distinguish between the pruned models.
We find that top5 loss is a better meta-attribute compared to top1 loss, as it reflects more information and is thus more general. The improvement of top5 meta-attributes over random meta-attributes validates the effectiveness of the meta pruning process.
From a statistical perspective, we also use the mean value of the network as a meta-attribute. The poor performance of mean value meta-attributes may be attributed to the fact that too much information about the network is lost in the mean calculation. 
Finding better meta-attributes is to be explored in future investigations.

%------------------------------------------------------

\begin{table}[!t]
\small
\setlength{\tabcolsep}{0.3em}
\begin{center}
% \color{red}\begin{tabular}{| c | c | c | c  | c |}
\begin{tabular}{| c | c | c | c  | c |}

\hline
\multirow{2}{*}{\shortstack{Meta\\ \RRRR{attribute} }} & \multirow{2}{*}{\shortstack{Top5}} & \multirow{2}{*}{\shortstack {Top1 }} & \multirow{2}{*}{\shortstack {Mean}} &\multirow{2}{*}{\shortstack {Random }}\\
               &         &          &        &          \\ \hline
Acc. (\%)     & 93.52$_{\pm0.56}$    &  93.39$_{\pm0.40}$  & 91.52$_{\pm0.34}$   &  91.82$_{\pm0.41}$   \\  \hline
\end{tabular}
\end{center}
\caption{
Accuracy of ResNet-110 on CIFAR-10 regarding different meta-attributes. The pruning rate is 40\%.
}
\label{table:mate-attr}
\end{table}

%------------------------------------------------------

\begin{figure}[!t]
\center
\includegraphics[width=0.7\linewidth]{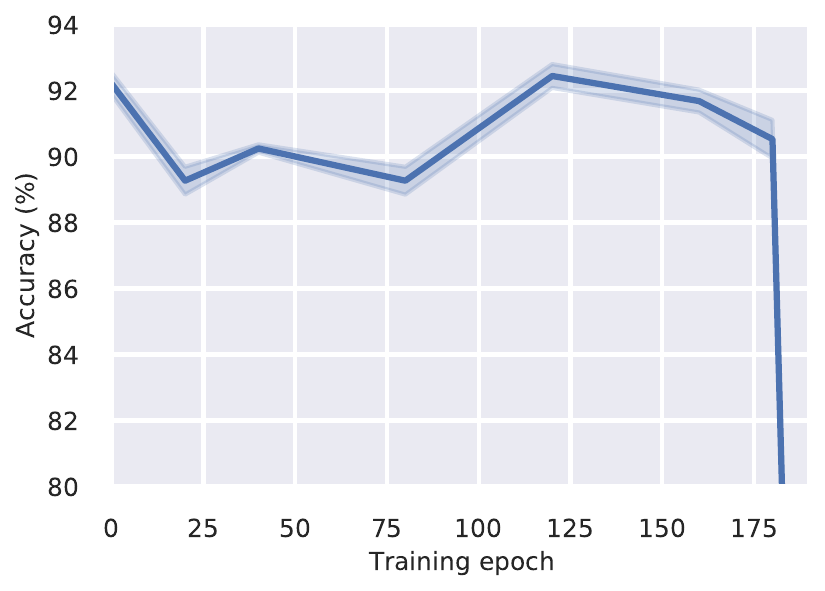}
\caption{
\RR{Varying training epochs to test the effects of the convergence of CNNs on the search results of \Eqref{eq:min-attribute}. The total training epoch for pruning ResNet-56 on CIFAR-10 is 200. (The solid line and shadow denote the mean and standard deviation of three experiments, respectively.)
}
}
\label{fig:train_epoch}
\end{figure}

\RR{\textbf{Varying Convergence.}}
\RR{
The state of convergence in CNNs influence the search results of \Eqref{eq:min-attribute}. To examine the convergence on the search results of \Eqref{eq:min-attribute}, we change the training epoch to test the final search results (In \Figref{fig:train_epoch}).
When the training epoch equals to zero, that is, we prune the model from scratch, the accuracy of the pruned model is about 92.5\%.
The reason why this setting could achieve good results might be due to the ``lottery ticket hypothesis'' proposed in \cite{frankle2018lottery}.
Specifically, randomly initialized networks contain subnetworks (``winning ticket'') that reach test accuracy comparable to the original network.
When the training epoch reaches 20 to 80, the accuracy drops to about 89\%.
This means the preliminary convergence at early epochs is not ideal to search for pruned models and may harm the accuracy.
Further increasing the training epoch to 120 would boost the accuracy.
The reason is that the convergence at later epochs is sufficient to search for a good pruned model.
If the training epoch continues to increase, the accuracy will drop dramatically.
This decrease in accuracy is because the network would not have enough training time to recover accuracy from the pruning operation. Since the total training epoch is 200, if the number of training epochs is larger than 180, the remaining epoch is smaller than 20, which is not enough for the network to recover accuracy.
}
 
%-------------------------------

\begin{figure}[!t]
\center
\includegraphics[width=0.7\linewidth]{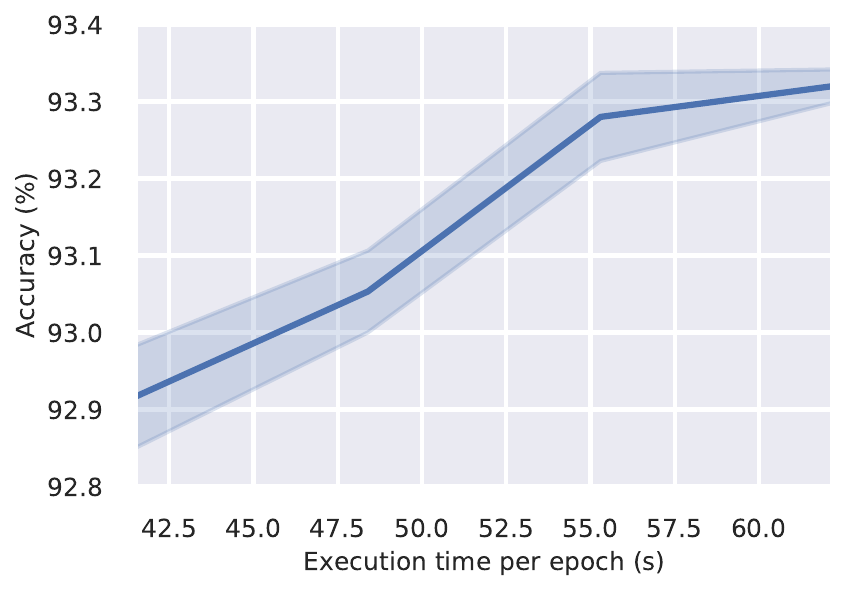}
\caption{
\RR{Varying execution time to test accuracies of pruned models. (The solid line and shadow denote the mean and standard deviation of three experiments, respectively.)
}
}
\label{fig:exc_time}
\end{figure}

\RR{\textbf{Varying Execution Time.}}
\RR{We use four neighboring CNNs in the experimental setting to balance the optimality and execution time. To better understand the influence of the execution time on the pruning results, we change the number of candidate neighboring CNNs so that the execution time could change. The results are shown in \Figref{fig:exc_time}. Generally, when we increase the number of neighboring CNNs from 2 to 5, the execution time would increase from 42 seconds to 63 seconds, and the accuracy would increase. This is because a larger search space is beneficial to the final accuracy. 
We also find that if execution time keeps increasing, the accuracy improvement will gradually become smaller. For example, increasing the execution time from 50 to 55 would lead to about 0.20\% accuracy improvement. In contrast, increasing the execution time from 57 to 62 would lead to only 0.03\% accuracy improvement. These results also demonstrate that our scope of search is appropriately defined. Further increasing the execution time or the scope of search does not bring considerable improvement.
% This means that Pearson correlation based criterion is not suitable for our pruning framework.
% When we add a new filter ranking criterion, that is, the Pearson correlation based criterion~\cite{benesty2009pearson}, into the search scope,
}

\subsection{\RR{Analysis of Pearson Correlation Based Criterion}}

\RRRR{We test the criterion that removes the filters that have a large Pearson correlation~\cite{benesty2009pearson} with other filters. It shows that our meta-process succeeds in excluding bad criteria during training.
}

\RRRR{
\textbf{Single Criterion.}
First, we compare the Pearson correlation-based method with the Minkowski distance-based criterion without meta-process, that is, only one criterion is utilized. The results are shown in \Tabref{table:pearson}. 
The Pearson correlation-based criterion is about 27\% less accurate than the Minkowski distance-based criterion when pruning the scratch models and has about 17\% less accuracy when pruning the pre-trained models. 
It is clear that the Pearson correlation-based criterion is not as effective as the Minkowski distance-based criterion to serve as a pruning criterion in our framework.
}

\begin{table}[!t]
\small
\setlength{\tabcolsep}{0.75em}
\begin{center}
% \color{red}\begin{tabular}{|c|c|c |c|}
\begin{tabular}{|c|c|c |c|}

\hline
Criteria  & pretrain? & Accuracy (\%)       \\ \hline
{Minkowski }                     & \xmark     & \textbf{92.37} ($\pm$0.30) \\\hline
Pearson                          & \xmark     & {65.01} ($\pm$10.07) \\\hline

Minkowski                        & \cmark     & \textbf{93.37} ($\pm$0.13) \\\hline
Pearson                          & \cmark     & {75.19} ($\pm$11.87) \\\hline

\end{tabular}
\end{center}
\caption{\RR{Pruning scratch and pre-trained ResNet-56 on CIFAR-10 with different criteria. }
}
\label{table:pearson}
\end{table}
%------------------------------------------------------

\RRRR{
\textbf{Multiple Criteria.}
In order to see whether the Pearson correlation-based criterion will be chosen in the meta-process in \Eqref{eq:min-attribute}, we add the Pearson correlation-based criterion as an additional candidate pruning criterion in the meta-process. The other four criteria are the same in the experiment settings in \Secref{sec:exp-pruning}. After examining the meta-process during the whole training process, we find that the Pearson correlation-based criterion is not selected. This result also shows that 
the Pearson correlation-based criterion is not an essential pruning criterion in our framework.
}

\begin{figure}[!t]
\center
\includegraphics[width=0.8\linewidth]{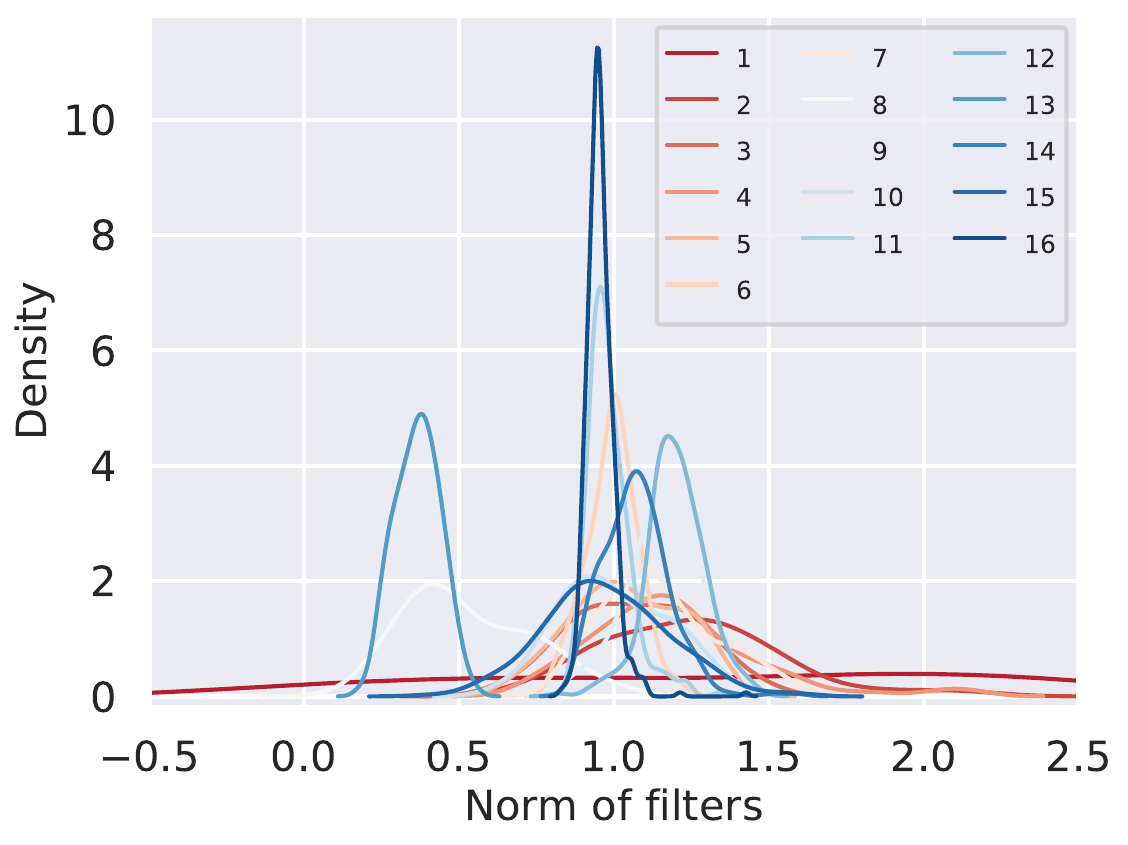}
\caption{
\RR{Norm distribution of filters from different layers of ResNet-18 on ILSVRC-2012. Different colors represent different layers. The density is obtained with Kernel Distribution Estimate (KDE) of the norm distribution.
}
}
\label{fig:pearson_exp}
\end{figure}

\RR{
\textbf{Explanation.}
To comprehensively understand the network, we plot the kernel distribution estimate (KDE)~\cite{silverman2018density} for the norm distribution of all convolutional layers, as shown in \Figref{fig:pearson_exp}. The norm distribution reflects the weight values of the filters. From the figure, we find that the lower layers have a broader range of norms than the upper layers. For example, the range of the norm of the first layer (indicated by the deep red line) is [0.5, 1.5], and the range of norm of the $16_{th}$ layer is [0.8, 1.1]. Since different layers have different functions~\cite{yosinski2015understanding}, these results imply that in neural networks, different functions are achieved by different filter values.
The above phenomenon may account for the Pearson correlation-based criterion is not a suitable pruning criterion. A key mathematical property of the Pearson correlation coefficient~\cite{benesty2009pearson} is that the Pearson correlation coefficient of two variables $X$ and $Y$ is invariant under changes in location and scale in these two variables. Specifically, transforming a variable $X$ to $a + bX$ would not change the Pearson correlation coefficient. Therefore, the Pearson correlation-based criterion could not distinguish different functions in the neural network, where the location and scale are crucial for conducting different functions.
}

\subsection{\RRRR{Analysis of Cosine Distance Based Criterion}}
\RRRR{
We also test the measure based on cosine distance~\cite{ye2011cosine} in \Tabref{table:cosine}. ``Only Cosine'' means just the cosine distance criterion is used as the pruning criterion, while ``Multiple'' indicates five criteria, including cosine distance criterion and four criteria listed in section IV. B.
We find that cosine distance criterion has low accuracy and is not suitable for pruning. Moreover, in the ``Multiple'' setting, we find the cosine distance criterion is never selected by our meta-process.
This means that our meta-process succeed in excluding the bad criterion, such as the cosine distance criterion, during the process of decision.
}

\begin{table}[!t]
\small
\setlength{\tabcolsep}{0.75em}
\begin{center}
% \color{red}\begin{tabular}{|c|c|c |c|}
\begin{tabular}{|c|c|c |c|}

\hline
Criteria  & pretrain? & Accuracy (\%)       \\ \hline
Only Cosine                     & \xmark     & {72.33} ($\pm$13.22) \\\hline
Multiple                      & \xmark     & {92.29} ($\pm$0.13) \\\hline

Only Cosine                        & \cmark     & {71.61} ($\pm$4.98) \\\hline
Multiple                          & \cmark     & \textbf{93.32} ($\pm$0.07) \\\hline

\end{tabular}
\end{center}
\caption{\RRRR{Pruning ResNet-56 on CIFAR-10 with cosine distance criterion. ``Multiple'' means five criteria including cosine distance criterion and four criteria listed in section IV. B.}
}
\label{table:cosine}
\end{table}
%------------------------------------------------------

\begin{table}[!t]
\small
\setlength{\tabcolsep}{0.75em}
\begin{center}
% \color{red}\begin{tabular}{|c|c|c |c|}
\begin{tabular}{|c|c|c |c|}

\hline
Weighted Criteria  & Selected Index & Accuracy (\%)       \\ \hline
Norm-1 + Dist-1                    & [3,13,4,15]     & \textbf{93.69} ($\pm$0.07) \\\hline
Norm-1 + Pearson                      &  [4,3,0,15]     & {92.25} ($\pm$0.68) \\\hline

Dist-1 + Pearson                     & [4,3,0,13]     & {92.26} ($\pm$1.26) \\\hline
Norm-1 + Dist-1 + Pearson                         & [4,3,13,15]     & {93.40} ($\pm$0.08) \\\hline

\end{tabular}
\end{center}
\caption{\RRRR{Pruning ResNet-56 on CIFAR-10 with different weighted criteria. ``Selected Index'' means the index of pruned filters in the first convolutional layer.}
}
\label{table:weight}
\end{table}
%------------------------------------------------------

\subsection{\RRRR{Analysis of Weighted Criteria}}

\RRRR{
We further examine the results of weighted criteria, and the results are shown in \Tabref{table:weight}. Specifically, we first obtain the rankings under each criterion. Then we calculate the average ranking for several candidate criteria. In this way, the average ranking can be viewed as the ranking of weighted criteria. We do not use the importance scores because the importance scores for different criteria has various ranges, which makes the weighting process difficult.
In \Tabref{table:weight}, ``Norm-1 + Dist-1'' means the weighted criteria for $\ell_{1}$-norm-based criterion and Manhattan distance-based criterion.
It is clear that ``Norm-1 + Dist-1'' achieves the best performance because both $\ell_{1}$-norm-based criterion and Manhattan distance-based criterion are good for pruning.
In contrast, other weighted criteria containing the Pearson criterion obtain smaller accuracy than ``Norm-1 + Dist-1''.
These results show that it is difficult for the weighted criteria to exclude the ``bad effects'' of bad criteria such as the Pearson criterion.
}

% , we propose a normalized weighting method 

\subsection{Feature Map Visualization and Explanation}
We visualize the feature maps of the first layer of the first block of ResNet-18, as shown in~\Figref{fig:visual}. We rank the 64 channel\footnote{The channels correspond the filters in the network.} in this layer with number 0 to 63 and set the pruning rate to 10\% to choose six filters to be pruned.
We select the channel (44, 29, 62, 56, 30, 51) via the $L2$-norm criterion and select the channel (44, 56, 29, 51, 52, 24) via Euclidean distance criterion. For both criteria, we select channel (29, 44, 51, 56), but order of the channels is different.

We focus on the different channels selected by these criteria to illustrate their difference.
In addition to the common part, the norm-based criterion select (30, 62), while the distance-based criterion select (24, 52).
Channel (30, 62) have rather small activation values and might be meaningless to the network, so they are selected by the norm-based criterion. For channel (24, 52), the rough shapes of the cat in these channels are similar to other channels such as (17, 23, 25, 32, 59), so the distance-based criterion (relational criterion) prefers to prune these channels.
These results validate our points that norm-based criterion and distance-based criterion consider different aspects of the network. In this way, during the updating of the network and the filter distribution, adaptively selecting those criteria is necessary.

%%%%%%%%%%%%%%%%%%%%%%%%%%%%%%%%%%%%%%%%%%%%%%%%%%%%%%%%%%%%%%%%%%%%%%%%%%%%%%%%%%%%%%%%%%%%%w%%%%%%

\section{Conclusion and Future Work}
 
In this paper, we propose a new \RRR{Meta-attribute-based Filter Pruning} (MFP) strategy for deep CNNs acceleration. Unlike the existing norm-based criterion, MFP explicitly considers the geometric information between filters. Furthermore, as a meta-framework, MFP adaptively selects the suitable criteria during training to fit the current filter distribution.
Results show that MFP advances the state-of-the-art methods in several benchmarks.
In the future, we could consider utilizing different criteria for different layers of the network.
Even within a network layer, we could combine different criteria for filter pruning.
Besides, whether a better meta-attribute exists is yet to be explored.
Moreover, other parallel acceleration algorithms, such as low-precision weights, could be used as a complementary method to improve the performance further.

\section{ACKNOWLEDGEMENT}
This work was supported by the Australian Research Council (ARC) under Grant DP200100938.
%%%%%%%%%%%%%%%%%%%%%%%%%%%%%%%%%%%%%%%%%%%%%%%%%%%%%%%%%%%%%%%%%%%%%%%%%%%%%%%%%%%%%%%%%%%%%%%%%%%

% use section* for acknowledgment
%\section*{Acknowledgment}

%The authors would like to thank...

% Can use something like this to put references on a page
% by themselves when using endfloat and the captionsoff option.
\ifCLASSOPTIONcaptionsoff
  \newpage
\fi

% trigger a \newpage just before the given reference
% number - used to balance the columns on the last page
% adjust value as needed - may need to be readjusted if
% the document is modified later
%\IEEEtriggeratref{8}
% The "triggered" command can be changed if desired:
%\IEEEtriggercmd{\enlargethispage{-5in}}

% references section

% can use a bibliography generated by BibTeX as a .bbl file
% BibTeX documentation can be easily obtained at:
% http://mirror.ctan.org/biblio/bibtex/contrib/doc/
% The IEEEtran BibTeX style support page is at:
% http://www.michaelshell.org/tex/ieeetran/bibtex/
%\bibliographystyle{IEEEtran}
% argument is your BibTeX string definitions and bibliography database(s)
%\bibliography{IEEEabrv,../bib/paper}
%
% <OR> manually copy in the resultant .bbl file
% set second argument of \begin to the number of references
% (used to reserve space for the reference number labels box)
\bibliographystyle{IEEEtran}
% \bibliography{IEEEfull,ref}
\bibliography{IEEEabrv,ref}

%\begin{thebibliography}{1}

%\bibitem{IEEEhowto:kopka}
%H.~Kopka and P.~W. Daly, \emph{A Guide to \LaTeX}, 3rd~ed.\hskip 1em plus 0.5em minus 0.4em\relax Harlow, England: Addison-Wesley, 1999.

%\end{thebibliography}

% biography section
% 
% If you have an EPS/PDF photo (graphicx package needed) extra braces are
% needed around the contents of the optional argument to biography to prevent
% the LaTeX parser from getting confused when it sees the complicated
% \includegraphics command within an optional argument. (You could create
% your own custom macro containing the \includegraphics command to make things
% simpler here.)
%\begin{IEEEbiography}[{\includegraphics[width=1in,height=1.25in,clip,keepaspectratio]{mshell}}]{Michael Shell}
% or if you just want to reserve a space for a photo:

\begin{IEEEbiography}
[{\includegraphics[width=1in,height=1.25in,clip,keepaspectratio]{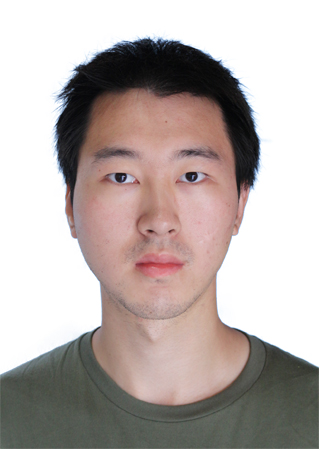}}]
{Yang He}
%Biography text here.
received the B.S. degree and MSc from the University of Science and Technology of China, Hefei, China, in 2014 and 2017, respectively. He is expected to obtain Ph.D. degree at the University of Technology Sydney in 2022. He is currently a scientist at A*STAR Centre for Frontier AI Research (CFAR), Singapore. His research interests include deep learning, computer vision, and filter pruning.
\end{IEEEbiography}

\begin{IEEEbiography}
[{\includegraphics[width=1in,height=1.25in,clip,keepaspectratio]{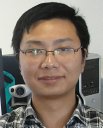}}]
{Ping Liu} is currently a scientist at A*STAR Centre for Frontier AI Research (CFAR), Singapore. Before joining CFAR, he was a research staff with the Center for Artificial Intelligence, University of Technology Sydney, Sydney, AUS. He received his Ph.D. degree in the Department of Computer Science and Engineering, University of South Carolina, SC, USA. He got his Master Degree from Huazhong University of Science and Technology, WuHan, China; Bachelor Degree from Wuhan University of Technology, WuHan, China. His research interests include computer vision, machine learning, and deep learning.
\end{IEEEbiography}

\begin{IEEEbiography}
[{\includegraphics[width=1in,height=1.25in,clip,keepaspectratio]{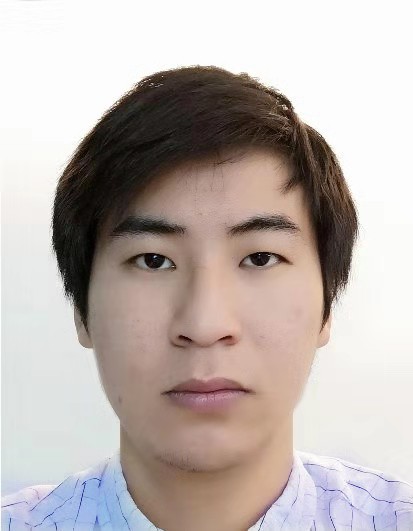}}]
{Linchao Zhu} is currently a Lecturer and core member of the Australian Artificial Intelligence Institute (AAII), University of Technology Sydney. He received his Ph.D. degree in University of Technology Sydney. He graduated from Zhejiang University with a bachelor’s degree. His research interests include video representation learning, unsupervised learning, self-supervised learning, few-shot learning, transfer learning, model efficiency.
\end{IEEEbiography}

\begin{IEEEbiography}
[{\includegraphics[width=1in,height=1.25in,clip,keepaspectratio]{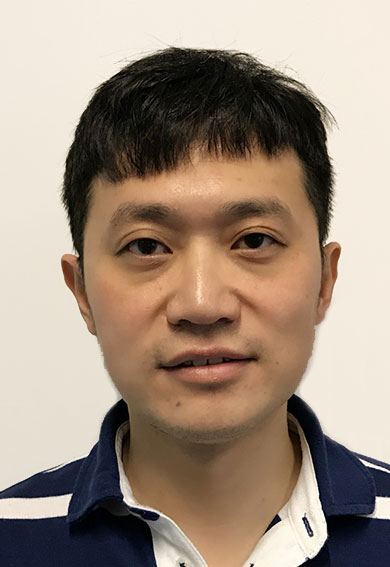}}]
{Yi Yang}
%Biography text here.
received the Ph.D. degree in computer science from Zhejiang University, Hangzhou, China, in 2010. He is currently a distinguished professor with Zhejiang University, China. 
Professor Yi Yang is an unremunerated Adjunct Professor with the Australian Artificial Intelligence Institute (AAII), University of Technology Sydney, Australia.
He was a professor and Director of the ReLER Lab at the Australian Artificial Intelligence Institute (AAII), University of Technology Sydney, Australia. He was a Post-Doctoral Research with the School of Computer Science, Carnegie Mellon University, Pittsburgh, PA, USA. His current research interest include machine learning and its applications to multimedia content analysis and computer vision, such as multimedia indexing and retrieval, surveillance video analysis and video semantics understanding.
\end{IEEEbiography}
% if you will not have a photo at all:
%\begin{IEEEbiographynophoto}{John Doe}
%Biography text here.
%\end{IEEEbiographynophoto}

% insert where needed to balance the two columns on the last page with
% biographies
%\newpage

% You can push biographies down or up by placing
% a \vfill before or after them. The appropriate
% use of \vfill depends on what kind of text is
% on the last page and whether or not the columns
% are being equalized.

%\vfill

% Can be used to pull up biographies so that the bottom of the last one
% is flush with the other column.
%\enlargethispage{-5in}

% that's all folks
\end{document}